\documentclass[11pt]{article}

\usepackage[numbers, compress]{natbib}

\usepackage[utf8]{inputenc} 
\usepackage[T1]{fontenc}    
\usepackage{url}            
\usepackage{booktabs}       
\usepackage{amsfonts}       
\usepackage{nicefrac}       
\usepackage{microtype}      

\usepackage{url}
\usepackage{smile}
\usepackage{algorithm}
\usepackage{algorithmic}
\usepackage{todonotes}
\usepackage{epstopdf}
\usepackage{wrapfig}
\usepackage{xcolor}
\usepackage[colorlinks, linkcolor={blue!50!black},
anchorcolor={blue!50!black}, citecolor={blue!50!black}]{hyperref}
\usepackage[margin=1in]{geometry}
\usepackage[normalem]{ulem}
\usepackage[export]{adjustbox}
\usepackage{mathtools, cuted}
\usepackage{enumitem}

\DeclareMathAlphabet{\mathsf}{OT1}{cmss}{m}{n}
\DeclarePairedDelimiter{\ceil}{\lceil}{\rceil}
\SetMathAlphabet{\mathsf}{bold}{OT1}{cmss}{bx}{n}

\usepackage{macros}

\def\shownotes{1}  
\ifnum\shownotes=1
\newcommand{\authnote}[2]{$\ll$\textsf{\small #1 notes: #2}$\gg$}
\else
\newcommand{\authnote}[2]{}
\fi

\title{\bf \huge Towards Understanding Hierarchical Learning: \\
  Benefits of Neural Representations}

\author{
  Minshuo Chen\thanks{Georgia Tech. \texttt{mchen393@gatech.edu}. Work
  done while at Salesforce Research.} \and
  Yu Bai\thanks{Salesforce Research. \texttt{yu.bai@salesforce.com}} \and
  Jason D. Lee\thanks{Princeton University. \texttt{jasonlee@princeton.edu}} \and
  Tuo Zhao\thanks{Georgia Tech. \texttt{tourzhao@gatech.edu}} \and
  Huan Wang\thanks{Salesforce
    Research. \texttt{\{huan.wang,cxiong,rsocher\}@salesforce.com}}
  \and
  Caiming Xiong\footnotemark[5] \and
  Richard Socher\footnotemark[5]
}

\newcommand{\commentout}[1]{}
\ifdefined\usebigfont

\usepackage{times}
\usepackage[fontsize=13pt]{scrextend}
\AtBeginDocument{
	\newgeometry{left=1.56in,right=1.56in,top=1.71in,bottom=1.77in}
}
\else
\fi

\begin{document}

\maketitle

\def\sec{Sections_arxiv}

\begin{abstract}

  Deep neural networks can empirically perform efficient hierarchical learning, in which the layers learn useful representations of the data. However, how they make use of the intermediate representations are not explained by recent theories that relate them to ``shallow learners'' such as kernels. In this work, we demonstrate that intermediate \emph{neural representations} add more flexibility to neural networks and can be advantageous over raw inputs. We consider a fixed, randomly initialized neural network as a representation function fed into another trainable network. When the trainable network is the quadratic Taylor model of a wide two-layer network, we show that neural representation can achieve improved sample complexities compared with the raw input: For learning a low-rank degree-$p$ polynomial ($p \geq 4$) in $d$ dimension, neural representation requires only $\widetilde{O}(d^{\ceil{p/2}})$ samples, while the best-known sample complexity upper bound for the raw input is $\widetilde{O}(d^{p-1})$. We contrast our result with a lower bound showing that neural representations do not improve over the raw input (in the infinite width limit), when the trainable network is instead a neural tangent kernel. Our results characterize when neural representations are beneficial, and may provide a new perspective on why depth is important in deep learning.
  
  
\end{abstract}
\ifdefined\minusspace\vspace{-0.15in}\else\fi
\section{Introduction}\label{sec:intro}
\ifdefined\minusspace\vspace{-0.1in}\else\fi

Deep neural networks have been empirically observed to be more powerful than their shallow counterparts on a variety of machine learning tasks~\cite{lecun2015deep}.
For example, on the ImageNet classification task, a $152$-layer residual network can achieve $8$\%-$10$\% better top-$1$ accuracy than a shallower $18$-layer ResNet~\cite{he2016deep}. A widely held belief on why depth helps is that deep neural networks are able to perform efficient \emph{hierarchical learning}, in which the layers learn representations that are increasingly useful for the present task. Such a hierarchical learning ability has been further leveraged in transfer learning. For example, \cite{girshick2014rich} and \cite{devlin2018bert} show that by combining with additional task-specific layers, the bottom layers of pre-trained neural networks for image classification and language modeling can be naturally transferred to other related tasks and achieve significantly improved performance.



Despite significant empirical evidence, we are in the lack of practical theory for understanding the hierarchical learning abilities of deep neural networks. Classical approximation theory has established a line of ``depth separation'' results which show that deep networks are able to approximate certain functions with much fewer parameters than shallow networks \cite{delalleau2011shallow, telgarsky2016benefits, eldan2016power,yarotsky2017error,chen2019efficient}. These work often manipulates the network parameters in potentially pathological ways, and it is unclear whether the resulting networks can be efficiently found through gradient-based optimization. A more recent line of work shows that overparametrized deep networks can be provably optimized and generalize as well as the so-called Neural Tangent Kernels (NTKs) \cite{jacot2018neural,du2018gradient,du2018gradient2,allen2018convergence,allen2019learning,arora2019fine}. However, these results do not take the hierarchical structure of the neural networks into account, and cannot justify any advantage of deep architectures. More recently, \cite{huang2020deep} show that some NTK models of deep networks are actually degenerate, and their generalization performance are no better than those associated with shallow networks.



In this paper, we provide a new persepctive for understanding hierarchical learning through studying intermediate \emph{neural representations}---that is, feeding fixed, randomly initialized neural networks as a representation function (feature map) into another trainable model. The prototypical model we consider is a wide two-layer neural network taking a representation function $\hb$ as the input, that is,
\begin{align}
  \label{equation:prototypical-model}
  f_\Wb(\xb) \defeq \frac{1}{\sqrt{m}} \sum_{r=1}^m a_r \phi(\wb_r^\top \hb(\xb)),
\end{align}
where $\xb\in\R^d$ is the feature, $\hb:\R^d\to\R^D$ is a data-independent representation function that is held fixed during learning, and $\Wb = [\wb_1, \dots, \wb_m]^\top \in\R^{m\times D}$ is the weight matrix to be learned from the data. For example, when $\hb(\xb)=\sigma(\Vb\xb+\bbb)$ is another one-hidden-layer network (i.e. neural representations), the model $f$ is a three-layer network in which we only learn the weight matrix $\Wb$. Studying this model will reveal how the lower-level representation affects learning in a three-layer network, a previously missing yet important aspect of hierarchical learning.



To demonstrate the importance of the representation function $\hb$, we investigate the \emph{sample complexity} for learning certain target functions using model~\eqref{equation:prototypical-model}. This is a fine-grained measure of the power of $\hb$ compared with other notions such as approximation ability. Indeed, we expect $f_\Wb$ to be able to approximate any ``regular'' (e.g. Lipschitz) function of $\xb$, whenever we use a non-degenerate $\hb$ and a sufficiently large width $m$. However, different choices of $\hb$ can result in different ways (for the trainable two-layer network) to approximate the same target function, thereby leading to different sample complexity guarantees. We will specifically focus on understanding when learning with the neural representation $\hb(\xb)=\sigma(\Vb\xb+\bbb)$ is more sample efficient than learning with the raw input $\hb(\xb)=\xb$, which is a sensible baseline for capturing the benefits of representations.


As the optimization and generalization properties of a general two-layer network can be rather elusive, we consider more \emph{optimization aware} versions of the prototype~\eqref{equation:prototypical-model}---we replace the trainable two-layer network in $f_\Wb$ by tractable alternatives such as its {\bf linearized model}~\cite{du2018gradient} (also known as ``lazy training'' in \cite{chizat2019lazy}) or {\bf quadratic Taylor model}~\cite{bai2019beyond}:
\leqnomode                  
\begin{align}
  & \tag*{({\tt NTK-}$\hb$)} f^L_\Wb(\xb) = \frac{1}{\sqrt{m}} \sum_{r=1}^m a_{r} \phi'(\wb_{0,r}^\top\hb(\xb)) (\wb_r^\top\hb(\xb)), \\
  & \tag*{({\tt Quad-}$\hb$)}  f^Q_\Wb(\xb) = \frac{1}{2\sqrt{m}} \sum_{r=1}^m a_{r} \phi''(\wb_{0,r}^\top\hb(\xb)) (\wb_r^\top\hb(\xb))^2 .
\end{align}
\reqnomode
When $\hb$ is the raw input (\ntkraw, \quadraw), these are models with concrete convergence and generalization guarantees, and can approximate the training of the full two-layer network in appropriate infinite-width limits (e.g.~\cite{du2018gradient,arora2019fine,allen2019learning,lee2019wide,bai2019beyond}). However, for learning with other representation functions, these models are less understood. The goal of this paper is to provide a quantitative understanding of these models, in particular when $\hb$ is a one-hidden-layer neural network (\ntkneural, \quadneural), in terms of their convergence, generalization, and sample complexities of learning.


%

The contributions of this paper are summarized as follows:
\begin{itemize}[wide]
\item We show that the \quadh~model has a benign optimization landscape, and prove generalization error bounds with a precise dependence on the norm of the features and weight matrices, as well as the conditioning of the empirical covariance matrix of the features (Section~\ref{section:opt-gen}).

\item We study sample complexities of learning when the representation is chosen as a one-hidden-layer neural network (\quadneural~model, Section~\ref{sec:upperbound}). For achieving a small excess risk against a low-rank degree-$p$ polynomial, we show that the \quadneural~model requires $\wt{O}(d^{\ceil{p/2}})$ samples. When $p$ is large, this is significantly better than the best known $\wt{O}(d^{p-1})$ upper bound for the \quadraw~model, demonstrating the benefits of neural representations.


\item When the trainable network is instead a linearized model (or an NTK), we present a lower bound showing that neural representations are provably \emph{not} beneficial: in a certain infinite-width limit, the \ntkneural~model requires at least $\Omega(d^p)$ samples for learning a degree-$p$ polynomial (Section~\ref{section:lower-bound}). Since $O(d^p)$ samples also suffice for learning with the \ntkraw~model, this shows that neural representations are not beneficial when fed into a linearized neural network.
\end{itemize}

\paragraph{Additional paper organization} We present the problem setup and algorithms in Section~\ref{section:prelim}, review related work in Section~\ref{section:related-work}, and provide conclusions as well as acknowledgments in Section~\ref{section:conclusion}.


\paragraph{Notations}
We use bold lower-case letters to denote vectors, e.g., $\xb \in \RR^{d}$, and bold upper-case letters to denote matrices, e.g., $\Wb \in \RR^{d_1 \times d_2}$. Given a matrix $\Wb \in \RR^{d_1 \times d_2}$, we let $\opnorm{\Wb}$ denote its operator norm, and $\norm{\Wb}_{2,4}$ denote its $(2, 4)$-norm defined as $\norm{\Wb}_{2, 4}^4 = \sum_{i=1}^{d_1} \norm{\Wb_{i, :}}_2^4$, where $\Wb_{i, :}\in\RR^{d_1}$ is the $i$-th row of $\Wb$. Given a function $f(\xb)$ defined on domain $\cX$ with a probability measure $\cD$, the $L_2$ norm is defined as $\norm{f}_{L_2}^2 = \int_{\cX} f^2(\xb) \cD(d\xb)$.

%
\ifdefined\minusspace\vspace{-0.15in}\else\fi
\section{Preliminaries}
\label{section:prelim}
\ifdefined\minusspace\vspace{-0.1in}\else\fi

\paragraph{Problem setup}
We consider the standard supervised learning task, in which we receive $n$ i.i.d. training samples $S_n = \{(\xb_i, y_i)\}_{i=1}^n$ from some data distribution $\cD$, where $\xb \in \cX$ is the input and $y \in \cY$ is the label. In this paper, we assume that $\cX = \mathbb{S}^{d-1} \subset \RR^d$ (the unit sphere) so that inputs have unit norm $\ltwo{\xb}=1$. Our goal is to find a predictor $f: \cX\mapsto \R$ such that the population risk
$$ \cR(f) \defeq \EE_{(\xb, y) \sim \cD} [\ell(f(\xb), y)] $$
is low, where $\ell:\RR\times \cY\to\RR$ is a loss function. We assume that $\ell(\cdot, y)$ is convex, twice differentiable with the first and second derivatives bounded by 1, and satisfies $|\ell(0, y)| \le 1$ for any $y\in \cY$. These assumptions are standard and are satisfied by commonly used loss functions such as the logistic loss and soft hinge loss.

Given dataset $S_n$, we define the empirical risk of a predictor $f$ as
\begin{align*}
  \hat{\cR}(f) \defeq \frac{1}{n} \sum_{i=1}^n \ell(f(\xb), y).
\end{align*}

\paragraph{Model, regularization, and representation}
We consider the case where $f$ is either the linearized or the quadratic Taylor model of a wide two-layer network that takes a fixed representation function as the input:
\leqnomode
\begin{align}
  & \tag*{(\ntkh)} f^L_\Wb(\xb) = \frac{1}{\sqrt{m}} \sum_{r=1}^m a_{r} \phi'(\wb_{0,r}^\top\hb(\xb)) (\wb_r^\top\hb(\xb)), \label{equation:ntkh} \\
  & \tag*{(\quadh)}  f^Q_\Wb(\xb) = \frac{1}{2\sqrt{m}} \sum_{r=1}^m a_{r} \phi''(\wb_{0,r}^\top\hb(\xb)) (\wb_r^\top\hb(\xb))^2, \label{equation:quadh}
\end{align}
\reqnomode
where $\hb:\R^d\to\R^D$ is a fixed representation function, $\wb_{0,r}\simiid \normal(\boldsymbol{0}, \Ib_D)$ and $a_{r}\simiid {\rm Unif}(\set{\pm 1})$ are randomly initialized and held fixed during the training, $\Wb = [\wb_1, \dots, \wb_m]^\top\in\R^{m\times D}$ is the trainable weight matrix\footnote{Our parameterization decouples the weight matrix in a standard two-layer network into two parts: the initialization $\Wb_0\in\R^{m\times D}$ that is held fixed during training, and the ``weight movement matrix'' $\Wb\in\R^{m\times D}$ that can be thought of as initialized at $\zero$.}, and $\phi:\R\to\R$ is a nonlinear activation. These models are taken as proxies for a full two-layer network of the form $\frac{1}{\sqrt{m}}\ab^\top \phi((\Wb_0+\Wb)\hb(\xb))$, so as to enable better understandings of their optimization.


For the \quadh~model, we add a regularizer to the risk so as to encourage $\Wb$ to have low norm. We use the regularizer
$\tfnorm{\Wb}^4=\sum_{r=1}^m \ltwo{\wb_r}^4$, and consider minimizing the regularized empirical risk 
\begin{align}
  \label{equation:regularized-risk}
  \hat{\cR}_\lambda(f^Q_\Wb) \defeq \hat{\cR}(f^Q_\Wb) + \lambda\tfnorm{\Wb}^4 = \frac{1}{n} \sum_{i=1}^n \ell(f^Q_\Wb(\xb_i), y_i) + \lambda\tfnorm{\Wb}^4.
\end{align}

In the majority of this paper, we will focus on the case where $\hb(\xb)$ is a fixed, randomly initialized neural network with one hidden layer of the form $\sigma(\Vb\xb+\bbb)$, with certain pre-processing steps when necessary. However, before we make the concrete choices, we think of $\hb$ as a general function that maps the raw input space $\R^d$ into a feature space $\R^D$ without any additional assumptions.

\paragraph{Connection to a three-layer model}
It is worth noticing that when $\hb$ is indeed a neural network, say $\hb(\xb)=\sigma(\Vb\xb)$ (omitting bias for simplicity), our \ntkh~and \quadh~models are closely related to the Taylor expansion of a \emph{three-layer network}
\begin{align*}
  \wt{f}_{\Wb, \Vb}(\xb) = \frac{1}{\sqrt{m}} \ab^\top\phi((\Wb_0+\Wb) \sigma(\Vb\xb)).
\end{align*}
Indeed, the \{\ntkh, \quadh\} models correspond to the \{linear, quadratic\} Taylor expansion of the above network over $\Wb$, and is thus a part of the full Taylor expansion of the three-layer network. By studying these Taylor models, we gain understandings about how deep networks use its intermediate representation functions, which is lacking in existing work on Taylorized models.


\ifdefined\minusspace\vspace{-0.15in}\else\fi
\section{Quadratic model with representations}
\label{section:opt-gen}
\ifdefined\minusspace\vspace{-0.1in}\else\fi

We begin by studying the (non-convex) optimization landscape as well as the generalization properties of the model~\ref{equation:quadh}, providing insights on what can be a good representation $\hb$ for such a model.

\paragraph{Base case of $\hb(\xb)=\xb$: a brief review} When $\hb(\xb)=\xb$ is the raw input, model~\ref{equation:quadh} becomes
\leqnomode
\begin{align}
  \label{equation:quad-raw}
  \tag*{(\quadraw)} & 
  f^Q_\Wb(\xb) = \frac{1}{2\sqrt{m}} \sum_{r=1}^m a_{r} \phi''(\wb_{0,r}^\top\xb) (\wb_r^\top \xb)^2,
\end{align}
\reqnomode
which is the quadratic Taylor model of a wide two-layer neural network. This model is analyzed by~\citet{bai2019beyond} who show that (1)
the (regularized) risk $\hat{\cR}_\lambda(f_\Wb)$ enjoys a nice optimization landscape despite being non-convex, and (2) the generalization gap of the model $f^Q_\Wb$ is controlled by $\tfnorm{\Wb}$ as well as $\|\frac{1}{n}\sum_{i\in[n]}\xb_i\xb_i^\top\|_{\rm op}$. Building on these results,~\citep{bai2019beyond} show that learning low-rank 
polynomials with~\ref{equation:quad-raw} achieves a better sample complexity than with the NTK. Besides the theoretical investigation, \citep{bai2020taylorized} empirically show that \ref{equation:quad-raw} model
also approximates the training trajectories of standard neural networks better than the linearized model.

\paragraph{General case}
We analyze optimization landscape and establish generalization guarantees when $\hb$ is a general representation function, extending the results in~\citep{bai2019beyond}. We make the following assumption:
\begin{assumption}[Bounded representation and activation]
  \label{assumption_input}
  There exists a constant $B_h$ such that $\norm{\hb(\xb)}_2 \leq B_h$ almost surely for $(\xb, y)\sim \cD$. The activation $\phi''$ is uniformly bounded: $\sup_{t\in\R}|\phi''(t)|\le C$ for some absolute constant $C$.
\end{assumption}

\begin{theorem}[Optimization landscape and generalization of \quadh]\label{thm:opt-gen}
  Suppose Assumption~\ref{assumption_input} holds.
  \label{theorem:opt-gen}
  \begin{enumerate}[wide,label=(\arabic*)]
  \item (Optimization)
    Given any $\epsilon > 0$, $\tau =\Theta(1)$, and some radius $B_{w,\star}>0$, suppose the width $m\ge \wt{O}(B_h^4B_{w,\star}^4\epsilon^{-1})$ and we choose a proper regularization coefficient $\lambda>0$. Then any second-order stationary point \footnote{$\Wb$ is a second-order stationary point (SOSP) of a twice-differentiable loss $L(\Wb)$ if $\grad L(\Wb)=\zero$ and $\grad^2 L(\Wb) \succeq \zero$.} (SOSP) $\hat{\Wb}$ of the regularized risk $\hat{\cR}_\lambda(f^Q_\Wb)$ satisfies $\|\hat{\Wb}\|_{2, 4} \leq O(B_{w, \star})$, and achieves
    \begin{align*}
\hat{\cR}_{\lambda}(f^Q_{\hat{\Wb}}) \leq (1+\tau) \min_{\tfnorm{\Wb} \le B_{w,\star}} \hat{\cR}(f^Q_\Wb) + \epsilon. 
    \end{align*}
\item (Generalization) 
  For any radius $B_w>0$, we have with high probability (over $(\ab, \Wb_0)$) that
\begin{align*}
  \E_{(\xb_i, y_i)} \brac{ \sup_{\tfnorm{\Wb}\le B_w } \abs{\cR(f^Q_{\Wb}) - \hat{\cR}(f^Q_{\Wb})} } \leq \tilde{O}\left(\frac{B_{h}^2 B_{w}^2 M_{h, \textrm{op}}}{\sqrt{n}} + \frac{1}{\sqrt{n}}\right),
\end{align*}
where $M_{h, \textrm{op}}^2 = B_h^{-2} \EE_{\xb} \left[\opnorm{\frac{1}{n} \sum_{i=1}^n \hb(\xb_i) \hb(\xb_i)^\top}\right]$.
\end{enumerate}
\end{theorem}
\paragraph{Efficient optimization; role of feature isotropicity}
Theorem~\ref{theorem:opt-gen} has two main implications: (1) With a sufficiently large width, any SOSP of the regularized risk $\hat{\cR}(f^Q_\Wb)$ achieves risk close to the optimum in a certain norm ball, and has controlled norm itself. Therefore, escaping-saddle type algorithms such as noisy SGD~\citep{jin2019stochastic,lee2016gradient} that can efficiently find SOSPs can also efficiently find these near global minima. (2) The generalization gap is controlled by $M_{h, {\rm op}}$, which involves the operator norm of $\frac{1}{n}\sum_{i=1}^n\hb(\xb_i)\hb(\xb_i)^\top$. It is thus beneficial if our representation $\hb(\xb)$ is (approximately) \emph{isotropic}, so that $M_{h, {\rm op}}\asymp O(1/\sqrt{D})$, which is much lower than its naive upper bound 1. This will be a key insight for designing our neural representations in Section~\ref{sec:upperbound}. The proof of Theorem~\ref{theorem:opt-gen} can be found in Appendix~\ref{appendix:proof-opt-gen}.



\ifdefined\minusspace\vspace{-0.15in}\else\fi
\section{Learning with neural representations}
\label{sec:upperbound}
\ifdefined\minusspace\vspace{-0.1in}\else\fi
We now develop theories for learning with neural representations, where we choose $\hb$ to be a wide one-hidden-layer neural network.

\ifdefined\minusspace\vspace{-0.1in}\else\fi
\subsection{Neural representations}
\ifdefined\minusspace\vspace{-0.05in}\else\fi

We consider a fixed, randomly initialized one-hidden-layer neural network:
\begin{align}
  \label{equation:g}
  \gb(\xb) = \sigma(\Vb\xb + \bbb) = \left[\sigma(\vb_1^\top \xb + b_1), \dots, \sigma(\vb_D^\top \xb + b_D) \right]^\top \in \R^D,
\end{align}
where $\vb_i\simiid \normal(0, \Ib_d)$ and $b_i\simiid \normal(0, 1)$ are the weights. Throughout this section we will use the indicator activation $\sigma(t)=\indic{t\ge 0}$. We will also choose $\phi(t)={\rm relu}(t)^2/2$ so that $\phi''(t)=\indic{t\ge 0}$ as well.\footnote{We can use a non-smooth $\sigma$ since $(\Vb, \bbb)$ are not trained. Our results can be extended to the situation where $\sigma$ or $\phi''$ is the relu activation as well.}

We define the representation function $\hb(\xb)$ as the \emph{whitened} version of $\gb(\xb)$:
\begin{align}
  \label{equation:neural-h}
  \hb(\xb) = \hat{\bSigma}^{-1/2} \gb(\xb), \quad \textrm{where} \quad \hat{\bSigma} = \frac{1}{n_0} \sum_{i=1}^{n_0} \gb(\wt{\xb}_i) \gb(\wt{\xb}_i)^\top.
\end{align}
Above, $\hat{\bSigma}$ is an estimator of the population covariance matrix\footnote{Strictly speaking, $\bSigma$ is the second moment matrix of $\gb(\xb)$.} $\bSigma = \EE_{\xb}[\gb(\xb) \gb(\xb)^\top]\in\R^{D\times D}$, and $\set{\wt{\xb}_i}_{i\in[n_0]}\eqdef \wt{S}_{n_0}$ is an additional set of unlabeled training examples of size $n_0$ (or a split from the existing training data). Such a whitening step makes $\hb(\xb)$ more isotropic than the original features $\gb(\xb)$, which according to Theorem~\ref{theorem:opt-gen} item (2) reduces the sample complexity for achieving low test error. We will discuss this more in Section~\ref{section:sample-complexity}.

We summarize our overall learning algorithm (with the neural representation) in Algorithm~\ref{alg}.






\begin{algorithm}[h]
\caption{Learning with Neural Representations (\quadneural~method)}
\label{alg}
\begin{algorithmic}
  \STATE {\bf Input}: Labeled data $S_n$, unlabeled data $\wt{S}_{n_0}$, initializations $\Vb\in\R^{D\times d}$, $\bbb\in\R^D$, $\Wb_0\in\R^{m\times D}$, parameters $(\lambda, \epsilon)$.
  \STATE {\bf Step 1:} Construct model $f^Q_\Wb$ as
  \leqnomode
  \begin{align}
    \label{equation:quad-neural}
    \tag*{(\quadneural)} & \qquad\qquad\qquad\quad
    f^Q_\Wb(\xb) = \frac{1}{2\sqrt{m}}\sum_{r=1}^m a_r\phi''(\wb_{0,r}^\top \hb(\xb)) (\wb_r^\top\hb(\xb))^2,
  \end{align}
  \reqnomode
  where $\hb(\xb)=\hat{\bSigma}^{-1/2}\gb(\xb)$ is the neural representation~\eqref{equation:neural-h} (using $\wt{S}_{n_0}$ to estimate the covariance).
  \STATE {\bf Step 2:} Find a second-order stationary point $\hat{\Wb}$ of the regularized empirical risk (on the data $S_n$):
  \begin{align*}
    \hat{\cR}_{\lambda}(f^Q_{\Wb}) = \frac{1}{n} \sum_{i=1}^n \ell(f^Q_{\Wb}(\xb_i), y_i) + \lambda \tfnorm{\Wb}^4.
  \end{align*}
\end{algorithmic}
\end{algorithm}


\ifdefined\minusspace\vspace{-0.1in}\else\fi
\subsection{Learning low-rank polynomials with neural representations}
\label{section:sample-complexity}
\ifdefined\minusspace\vspace{-0.05in}\else\fi


We now study the sample complexity of Algorithm~\ref{alg} to achieve low excess test risk compared with the best \emph{low-rank degree-$p$ polynomial}, that is, sum of polynomials of the form $(\bbeta^\top\xb)^p$.
This setting has been considered in a variety of prior work on learning polynomials~\citep{mu2014square,chen2020learning} as well as analyses of wide neural networks~\citep{arora2019fine,bai2019beyond}.

We need the following additional assumption on the random features.
\begin{assumption}[Lower Bounded Covariance]
  \label{assumption_covariance}
  For any $k$ and $D\le O(d^k)$, with high probability over $\Vb,\bbb$ (as $d\to\infty$), we have the minimum eigenvalue $\lambda_{\min}(\bSigma) \geq \lambda_k$ for some constant $\lambda_k>0$ that only depends on $k$ but not $d$, where $\bSigma =
  \EE_{\xb} [\sigma\left(\Vb \xb + \bbb\right) \sigma\left(\Vb \xb + \bbb\right)^\top]$ .
\end{assumption}
Assumption~\ref{assumption_covariance} states the features $\set{\sigma(\vb_i^\top\xb+\bbb_i)}$ to be not too correlated, which roughly requires the distribution of $\xb$ to span all directions in $\R^d$. For example, when $\xb\sim{\rm Unif}(\SSS^{d-1})$ (and with our choice of $\sigma(t) = \indic{t\ge 0}$), we show that this assumption is satisfied with
\begin{align*}
  \lambda_k = \Theta\paren{ \min_{{\rm deg}(q) \le k}\E_{z\sim\normal(0,1)}[ (\sigma(z) - q(z))^2]} \asymp k^{-1/2},
\end{align*}
where $q(z)$ denotes a polynomial in $z$ and its degree is denoted as ${\rm deg}(q)$.
For general distributions of $\xb$, we show Assumption~\ref{assumption_covariance} still holds under certain moment conditions on the distribution of $\xb$ (see the formal statement and proof of both results in Appendix~\ref{appendix:covariance}).


\paragraph{Sample complexity for learning polynomials}
We focus on low-rank polynomials of the form
\begin{align}
  \label{equation:f-star}
  f_\star(\xb) = \sum_{s=1}^{r_\star} \alpha_s (\bbeta_s^\top \xb)^{p_s},~~~\textrm{where}~~~|\alpha_s|\le 1,~\norm{(\bbeta_s^\top\xb)^{p_s}}_{L_2}\le 1,~p_s\le p~~\textrm{for all}~s.
\end{align}
We state our main result for the \quadneural~model to achieve low excess risk over such functions.

\begin{theorem}[Sample complexity of learning with \quadneural]
\label{thm:quadneural_generalization}
Suppose Assumption \ref{assumption_covariance} holds, and there exists some $f_\star$ of the form~\eqref{equation:f-star} that achieves low risk: $\cR(f_\star)\le \opt$.
Then for any $\epsilon,\delta \in (0, 1)$ and $\tau = \Theta(1)$, choosing 
\begin{align}
  D = \Theta\paren{{\rm poly}(r_\star, p) \sum_s \ltwo{\bbeta_s}^{2\ceil{p_s/2}} \epsilon^{-2}\delta^{-1}},~~~~m \ge \wt{O}\paren{ {\rm poly}(r_\star, D)\epsilon^{-2}\delta^{-1}},
\end{align}
$n_0= \wt{O}(D\delta^{-2})$, and a proper $\lambda > 0$, Algorithm~\ref{alg} achieves the following guarantee:
with probability at least $1-\delta$ over the randomness of data and initialization, any second-order stationary point $\hat{\Wb}$ of $\hat{\cR}_\lambda(f^Q_\Wb)$ satisfies
\begin{align*}
  \cR(f^Q_{\hat{\Wb}}) \le (1+\tau)\opt + \hspace{-0.0in}\underbrace{\epsilon}_{\substack{\textrm{approx.,} \\\textrm{requires large}~D}}\hspace{-0.0in} + \underbrace{\wt{O}\paren{\sqrt{ \frac{{\rm poly}(r_\star, p, \delta^{-1}) \lambda_{\ceil{p/2}}^{-1} \epsilon^{-2}\sum_{s=1}^{r_\star} \ltwo{\bbeta_s}^{2\ceil{p_s/2}}}{n}}}}_{\textrm{generalization, requires large}~n~\textrm{(given}~\epsilon{\rm )}}.
\end{align*}
In particular, for any $\epsilon>0$, we can achieve $\cR(f^Q_{\hat{\Wb}}) \le (1+\tau)\opt + 2\epsilon$ with sample complexity
\begin{align}
  \label{equation:sample-complexity}
  n_0 + n \le \wt{O}\paren{ {\rm poly}(r_\star, p, \lambda_{\ceil{p/2}}^{-1}, \epsilon^{-1}, \delta^{-1}) \sum_{s=1}^{r_\star} \ltwo{\bbeta_s}^{2\ceil{p_s/2}} }.
\end{align}
\end{theorem}
According to Theorem~\ref{thm:quadneural_generalization}, \quadneural~can learn polynomials of any degree by doing the following: (1) Choose a sufficiently large $D$, so that the neural representations are expressive enough; (2) Choose a large width $m$ in the quadratic model so as to enable a nice optimization landscape, where such $m$ only appears logarithmically in generalization error (Theorem~\ref{theorem:opt-gen}).
  
\paragraph{Improved dimension dependence over \quadraw~and \ntkraw}
We parse the sample complexity bound in Theorem~\ref{thm:quadneural_generalization} in the following important case: $\xb$ is relatively uniform (e.g. ${\rm Unif}(\SSS^{d-1})$), $\ltwo{\bbeta_s}=O(\sqrt{d})$, and the data is noiseless and realized by $f_\star$ (so that $\opt=0$). In this case we have $\norm{(\bbeta_s^\top\xb)^{p_s}}_{L_2}=O(1)$, and Assumption~\ref{assumption_covariance} holds with $\lambda_{\ceil{p/2}} \ge p^{-O(1)}$.
Thus, when only highlighting the $d$ dependence\footnote{For example, in the high-dimensional setting when $\epsilon=\Theta(1)$ and $d$ is large~\citep{ghorbani2019linearized}.},
the sample complexity required to achieve $\epsilon$ test risk with the \quadneural~is (reading from~\eqref{equation:sample-complexity})
\begin{align*}
  N_{\tt quad-neural} = \tilde{O}\paren{ d^{\ceil{p/2}} }.
\end{align*}
In comparison, the sample complexity for learning with the \quadraw~(quadratic neural network with the raw input) is
\begin{align*}
  N_{\tt quad-raw} = \tilde{O}\paren{d^{p-1}}
\end{align*}
(see, e.g.~\citep[Thm 7]{bai2019beyond}). 
Therefore, Theorem~\ref{thm:quadneural_generalization} shows that neural representations can significantly improve the sample complexity over the raw input, when fed into a quadratic Taylor model. 

\paragraph{Overview of techniques}
At a high level, the improved sample complexity achieved in Theorem~\ref{thm:quadneural_generalization} 
is due to the flexibility of the neural representation: the \quadh~model can express polynomials \emph{hierarchically}, using weight matrices with much smaller norms than that of a shallow learner such as the \quadraw~model. This lower norm in turn translates to a better generalization bound (according to Theorem~\ref{thm:opt-gen}) and an improved sample complexity. We sketch the main arguments here, and leave the complete proof to Appendix~\ref{pf:upperbound}.





\begin{enumerate}[label=(\arabic*),wide]
\item {\bf Expressing functions using hierarchical structure}: We prove the existence of some $\Wb^*\in\R^{m\times D}$ such that $f^Q_{\Wb^*}\approx f_\star$ by showing the following: (1) As soon as $D\ge \wt{O}(d^k)$, $\hb(\xb)$ can linearly express certain degree-$k$ polynomials as ``bases''; (2) For large $m$, the top quadratic taylor model can further express degree $\ceil{p_s/k}$ polynomials of the bases, thereby expressing $f_\star$. This is an explicit way of utilizing the hierarchical structure of the model. We note that our proof used $k=\ceil{p/2}$, but the argument can be generalized to other $k$ as well. 
  
\item {\bf Making representations isotropic}: We used a whitened version of a one-hidden-layer network as our representation function $\hb$ (cf.~\eqref{equation:neural-h}). The whitening operation does not affect the expressivity argument in part (1) above, but helps improve the conditioning of the feature covariance matrix (cf. the quantity $M_{h, {\rm op}}$ in Theorem~\ref{theorem:opt-gen}). Applying whitening, we obtain nearly isotropic features: $\E_{\xb}[\hb(\xb)\hb(\xb)^\top]\approx \Ib_D$, which is key to the sample complexity gain over the \quadraw~model as discussed above. 
  We note that well-trained deep networks with BatchNorm may have been implicitly performing such whitening operations in practice~\citep{morcos2018on}. We also remark that the whitening step in Algorithm~\ref{alg} may be replaced with using unwhitened representations with a \emph{data-dependent} regularizer, e.g., $\sum_{r=1}^m \|\hat\bSigma^{1/2} \wb_r\|_2^4$, which achieves similar sample complexity guarantees (see Appendix \ref{pf:upperbound}).

\end{enumerate}

\ifdefined\minusspace\vspace{-0.15in}\else\fi
\section{NTK with neural representations: a lower bound}
\label{section:lower-bound}
\ifdefined\minusspace\vspace{-0.1in}\else\fi
In this section, we show that neural representations may not be
beneficial over raw inputs when the trainable network is a
\emph{linearized} neural network through presenting a sample
complexity lower bound for this method in the infinite width
limit. 



More concretely, we consider \ntkneural, which learns a model $f^L_\Wb$ of the form
\leqnomode
\begin{align}
  \label{equation:linearized-model}
  \tag*{(\ntkneural)} & 
  \qquad \qquad\qquad f^L_\Wb(\xb) \defeq \frac{1}{\sqrt{m}} \sum_{r=1}^m
  a_r\phi'\paren{\wb_{0,r}^\top \gb(\xb) / \sqrt{D}}
  \paren{\wb_r^\top \gb(\xb) / \sqrt{D}}, 
\end{align}
\reqnomode
where 
$\gb(\xb) \defeq [\sigma(\vb_1^\top \xb + b_1),
\dots, \sigma(\vb_D^\top \xb + b_D)]^\top\in \R^D$ are the neural
random features (same as in~\eqref{equation:g}), and the  $1/\sqrt{D}$ factor rescales $\gb(\xb)$ to $O(1)$ norm on average.



\paragraph{Infinite-width limit: a kernel predictor}
Model~\ref{equation:linearized-model} is linear model with parameter $\Wb$, and can be viewed as a kernel predictor with a (finite-dimensional kernel) $H_{m,D}:\mathbb{S}^{d-1}\times \mathbb{S}^{d-1}\to\R$.
In the infinite-width limit of $D,m\to\infty$, we have $H_{m,D}\to H_\infty$, where
\begin{align*}
  & H_\infty(\xb, \xb') \defeq \E_{(u,v)\sim \normal(\zero, \bSigma(\xb, \xb'))} [\phi'(u)\phi'(v)] \cdot \Sigma_{12} (\xb, \xb'),~{\rm and}\\
  & \bSigma(\xb, \xb') = \begin{pmatrix}
    \E_{\vb, b}[\sigma(\vb^\top \xb + b)^2] & \E_{\vb, b}[\sigma(\vb^\top \xb+b)\sigma(\vb^\top \xb'+b)] \\
    \E_{\vb, b}[\sigma(\vb^\top \xb+b)\sigma(\vb^\top \xb'+b)] & \E_{\vb, b}[\sigma(\vb^\top \xb' + b)^2]
  \end{pmatrix},
\end{align*}
(see e.g.~\citep{jacot2018neural,du2019gradient} for the derivation). 
Motivated by this, we consider kernel predictors of the form
\begin{align}
  \label{equation:exact-kernel-predictor}
  \hat{f}_\lambda = \argmin_{f} \sum_{i=1}^n \ell(f(\xb_i), y_i) + \lambda\norm{f}_{H_\infty}^2
\end{align}
as a proxy for~\ref{equation:linearized-model},
where $\norm{\cdot}_{H_\infty}^2$ denotes the RKHS (Reproducing Kernel Hilbert Space) norm associated with kernel $H_\infty$. This set of predictors is a reliable proxy for the \ref{equation:linearized-model}~method: for example, taking $\lambda\to 0_+$, it recovers the solution found by gradient descent (with a small stepsize) on the top layer of a wide three-layer network~\citep{du2019gradient}.

We now present a lower bound for the predictor $\hat{f}_\lambda$, adapted from~\citep[Theorem 3]{ghorbani2019linearized}.
\begin{theorem}[Lower bound for \ntkneural]
  \label{theorem:lower-bound}
  Suppose the input distribution is $\xb\sim{\rm Unif}(\mathbb{S}^{d-1})$, and $y_\star= f_\star(\xb)$ where $f_\star\in L_2({\rm Unif}(\mathbb{S}^{d-1}))$ consists of polynomials of degree at least $p$\footnote{That is, $\|{\sf P}_{<p}f_\star\|_{L_2}=0$, where ${\sf P}_{<p}$ denotes the $L_2$ projection onto the space of degree $<p$ polynomials.}. Assume the sample size $n\le O(d^{p-\delta})$ for some $\delta>0$. Then for any fixed $\epsilon\in(0,1)$, as $d\to\infty$, the predictor $\what{f}_\lambda$ defined in~\eqref{equation:exact-kernel-predictor} suffers from the following lower bound with high probability (over $\set{(\xb_i, y_i)}$):
  \begin{align*}
    \E_{\xb} \brac{ \inf_{\lambda>0} (\hat{f}_\lambda(\xb) - f_\star(\xb))^2} \ge (1-\epsilon) \E_{\xb}[f_\star(\xb)]^2,
  \end{align*}
  that is, any predictor of the form~\eqref{equation:exact-kernel-predictor} will not perform much better than the trivial zero predictor.
\end{theorem}

\paragraph{No improvement over \ntkraw; benefits of neural representations}
Theorem~\ref{theorem:lower-bound} shows that the infinite width version~\eqref{equation:exact-kernel-predictor} of the \ntkneural~method requires roughly at least $\Omega(d^p)$ samples in order to learn any degree-$p$ polynomial up to a non-trivial accuracy (in squared error).
Crucially, this lower bound implies that \ntkneural~\emph{does not} improve over \ntkraw~(i.e. NTK with the raw input) in the infinite width limit---the infinite width \ntkraw~already achieves sample complexity \emph{upper bound} of $O(d^p)$ for learning a degree-$p$ polynomial $y=f_\star(\xb)$ when $\xb\sim{\rm Unif}(\mathbb{S}^{d-1})$~\citep{ghorbani2019linearized}. This is in stark contrast with our Theorem~\ref{thm:quadneural_generalization} which shows that \quadneural~improves over \quadraw, suggesting that neural representations are perhaps only beneficial when fed into a sufficiently complex model.




\ifdefined\minusspace\vspace{-0.15in}\else\fi
\section{Related work}
\label{section:related-work}
\ifdefined\minusspace\vspace{-0.1in}\else\fi

{\bf Approximation theory and depth separation.} Extensive efforts have been made on the expressivity of neural networks and the benefits of increased depth. Two separate focuses were pursued: 1) Universal approximation theory for approximating dense function classes, e.g., Sobolev and squared integrable functions \citep{cybenko1989approximation, hornik1991approximation, barron1993universal, irie1988capabilities, funahashi1989approximate, chui1992approximation, leshno1993multilayer, mhaskar1996neural, makovoz1996random}; 2) depth separation theory demonstrating the benefits of increased depth on expressing certain structured functions, e.g., saw-tooth functions \citep{haastad1987computational, delalleau2011shallow, rossman2015average, telgarsky2016benefits, eldan2016power}.
More recently, the recent work \citep{yarotsky2017error} merged the two focuses by studying unbounded-depth ReLU networks for approximating Sobolev functions.
In all these work, the network parameters are constructed in potentially weird ways, and it is unclear whether such networks can be efficiently found using gradient-based optimization.


\paragraph{Neural tangent kernels and beyond}
A growing body of recent work show the connection between gradient descent on the full network and the Neural Tangent Kernel (NTK)~\citep{jacot2018neural}, from which one can prove concrete results about neural network training~\citep{li2018learning,du2018gradient, du2019gradient,allen2018convergence,zou2018stochastic} and generalization~\citep{arora2019fine,allen2019learning,cao2019generalization}. 
Despite such connections, these results only show that neural networks are as powerful as shallow learners such as kernels. The gap between such shallow learners and the full neural network has been established in theory by~\citep{wei2019regularization,allen2019can,yehudai2019power,ghorbani2019limitations,woodworth2020kernel,dyer2019asymptotics} and observed in practice~\citep{arora2019exact,lee2019wide,chizat2018note}. Higher-order expansions of the \{network, training dynamics\} such as Taylorized Training~\citep{bai2019beyond,bai2020taylorized} and the Neural Tangent Hierarchy~\citep{huang2019dynamics} have been recently proposed towards closing this gap. Finally, recent work by~\citet{allen2020backward} shows that there exists a class of polynomials that can be efficiently learned by a deep network but not any ``non-hiearchical'' learners such as kernel methods or neural tangent kernels, thereby sheding light on how representations are learned hierarchically.

\paragraph{Learning low-rank polynomials in high dimension}
In \cite{mu2014square} and \cite{montanari2018spectral}, the authors propose a tensor unfolding algorithm to estimate a rank $k$ order $p$ tensor with $(d)^{p/2}k$ samples. Under Gaussian input data, \cite{chen2020learning} propose a Grassmanian manifold optimization algorithm with spectral initialization to estimate a polynomial over $k$-dimensional subspace of variables of degree $p$ with $O_{k,p}( d \log^d p)$ samples, where $O_{k,p}$ suppresses unknown (super)-exponential dependence on $k$ and $p$. However, these methods explicitly use knowledge about the data distribution.
Neural networks can often learn polynomials in distribution-free ways.
\citep{andoni2014learning,arora2019fine} show that wide two-layer networks that simulate an NTK require $\wt{O}(d^p)$ samples to learn a degree-$p$ polynomial. \citep{ghorbani2019linearized} show that $\Omega(d^p)$ samples is also asymptotically necessary for any rotationally invariant kernel. \citep{bai2019beyond} show that a randomized wide two-layer network requires $\wt{O}(d^{p-1})$ samples instead by coupling it with the quadratic Taylor model. Our algorithm belongs to this class of distribution-free methods, but achieve an improved sample complexity when the distribution satisfies a mild condition.




\ifdefined\minusspace\vspace{-0.15in}\else\fi
\section{Conclusion}
\label{section:conclusion}
\ifdefined\minusspace\vspace{-0.1in}\else\fi
This paper provides theoretical results on the benefits of neural representations in deep learning. We show that using a neural network as a representation function can achieve improved sample complexity over the raw input in a neural quadratic model, and also show such a gain is not present if the model is instead linearized. We believe these results provide new understandings to hiearchical learning in deep neural networks. For future work, it would be of interest to study whether deeper representation functions are even more beneficial than shallower ones, or what happens when the representation is fine-tuned together with the trainable network.

\section*{Acknowledgment}
We thank the anonymous reviewers for the suggestions. We thank Song Mei for the discussions about the concentration of long-tailed covariance matrices. JDL acknowledges support of the ARO under MURI Award W911NF-11-1-0303, the Sloan Research Fellowship, and NSF CCF 2002272. 

\bibliographystyle{abbrvnat}
\bibliography{ref,large_lr,appendix}

\appendix


\newpage
\section{Proofs for Section~\ref{section:opt-gen}}
\label{appendix:proof-opt-gen}
\subsection{Proof of Optimization in Theorem \ref{thm:opt-gen}}
We first derive the gradient and Hessian of empirical risk $\hat\cR(f^Q_\Wb)$, which will be used throughout the rest of the proof. For a better presentation, we denote $\inner{\cdot}{\cdot}$ as inner product and
\begin{align*}
f^Q_{\Wb}(\xb) = \frac{1}{2\sqrt{m}} \inner{\xb\xb^\top}{\Wb \Db(\xb) \Wb^\top} \text{ for }\Db_{rr}(\xb) = a_{r} \phi''(\wb_{0, r}^\top \hb(\xb)).
\end{align*}
We compute the gradient and Hessian of $\hat{\cR}(f^Q_{\Wb})$ along a given direction $\Wb_\star$.
\begin{align*}
\nabla_{\Wb} \hat{\cR}(f^Q_\Wb) & = \frac{2}{n}\sum_{i=1}^n \ell'( f^Q_\Wb(\xb_i), y_i) \frac{1}{2\sqrt{m}}\xb_i\xb_i^\top\Wb\Db(\xb_i) \quad \textrm{and} \\
\nabla^2_{\Wb} \hat\cR(f^Q_\Wb)[\Wb_\star, \Wb_\star] & = \frac{2}{n}\sum_{i=1}^n \ell'(f^Q_\Wb(\xb_i), y_i) \cdot \underbrace{\frac{1}{2\sqrt{m}}\inner{\xb_i\xb_i^\top}{\Wb_\star \Db(\xb_i) \Wb_\star^\top}}_{f^Q_{\Wb_\star}(\xb_i)} \\ 
& \quad+ 
       \frac{4}{n}\sum_{i=1}^n \ell''(f^Q_\Wb(\xb_i), y_i) \cdot
       \Bigg(\underbrace{\frac{1}{2\sqrt{m}}
       \inner{\xb_i\xb_i^\top}{\Wb\Db(\xb_i)\Wb_\star^\top}}_{\tilde{y}_i}\Bigg)^2 \\
& = \underbrace{\frac{2}{n} \sum_{i=1}^n \ell'(f^Q_\Wb(\xb_i), y_i) f^Q_{\Wb_\star}(\xb_i)}_{{\rm I}} + \underbrace{\frac{4}{n} \sum_{i=1}^n \ell''(f^Q_\Wb(\xb_i), y_i)\tilde{y}_i^2}_{{\rm II}}.  
\end{align*}
We denote $\hat{\cD}$ as the empirical data distribution, and bound I and II separately. 
\begin{align*}
{\rm I} & = 2\E_{\hat{\cD}}[\ell'(f^Q_\Wb(\xb), y)
    f^Q_{\Wb_\star}(\xb)] \\
  & = 2\E_{\hat{\cD}}[\ell'(f^Q_\Wb(\xb), y)
    f^Q_\Wb(\xb)] + 2\E_{\hat{\cD}}[\ell'(f^Q_\Wb(\xb), y)
    (f^Q_{\Wb_\star}(\xb) - f^Q_{\Wb}(\xb))] \\
  & \stackrel{(i)}{\le} \inner{\grad \hat\cR(f^Q_\Wb)}{\Wb} + 2\E_{\hat{\cD}}[\ell(
    f^Q_{\Wb_\star}(\xb), y) -
    \ell(f^Q_\Wb(\xb), y)] \\
  & = \inner{\nabla \hat\cR(f^Q_\Wb)}{\Wb} - 2(\hat\cR(f^Q_\Wb) - \hat\cR(f^Q_{\Wb^*})),
\end{align*}
where (i) follows directly by computing $\inner{\nabla \hat\cR(f^Q_\Wb)}{\Wb}$ and
the convexity of $\ell$.
For II, with $\ell'' \leq 1$, we have
\begin{align*}
{\rm II} \leq \frac{4}{n} \sum_{i=1}^n \sum_{r\leq m} \tilde{y}_i^2 & = \EE_{\hat\cD} \sbr{\frac{2}{m}\sum_{r\le m} 
    \phi''(\wb_{0,r}^\top\hb(\xb))^2(\wb_r^\top\hb(\xb))^2
    (\wb_{\star,r}^\top\hb(\xb))^2} \\
  & \le C^2 \EE_{\hat\cD}\sbr{\frac{1}{m}\sum_{r\le m}
    (\wb_r^\top\hb(\xb))^2(\wb_{\star,r}^\top\hb(\xb))^2}
  \\
  & \le \frac{1}{m} C^2 B_h^4 \sum_{r\le m}\norm{\wb_r}_2^2\norm{\wb_{\star,r}}_2^2 \\
  &  \le m^{-1} C^2 B_h^4\norm{\Wb}_{2, 4}^2
    \norm{\Wb_\star}_{2, 4}^2, 
\end{align*}
where the last step used Cauchy-Schwarz on $\set{\ltwo{\wb_r}}$ and
$\set{\ltwo{\wb_{\star,r}}}$, and the constant $C$ is the uniform upper bound on $\phi''$. Putting terms I and II together, we have
\begin{align}\label{eq:negative-lambda-bound}
\nabla^2_{\Wb} \hat\cR(f^Q_\Wb)[\Wb_\star, \Wb_\star] \leq \inner{\nabla \hat\cR(f^Q_\Wb)}{\Wb} - 2(\hat\cR(f^Q_\Wb) - \hat\cR(f^Q_{\Wb^*})) + m^{-1} C^2 B_h^4\norm{\Wb}_{2, 4}^2 \norm{\Wb_\star}_{2, 4}^2.
\end{align}

\begin{proof}[Proof of Theorem \ref{thm:opt-gen}, Optimization Part]\label{pf:opt}
We denote $\Wb^* = \argmin_{\norm{\Wb}_{2, 4} \leq B_{w, \star}} \hat{\cR}(f^Q_\Wb)$ and let its risk $\hat{\cR}(f^Q_{\Wb^*}) = M$. We begin by choosing the regularization strength as
\begin{equation*}
  \lambda = \lambda_0 B_{w, \star}^{-4},
\end{equation*}
where $\lambda_0$ is a constant to be determined.

We argue that any second order stationary point $\hat\Wb$ has to
satisfy $\|\hat\Wb\|_{2, 4} = O(B_{w, \star})$. We have for
any $\Wb$ that
\begin{align*}
\inner{\nabla \hat\cR(f^Q_\Wb)}{\Wb} & = \EE_{\hat\cD}\sbr{\ell'(
    f^Q_{\Wb}(\xb), y) \cdot 2f^Q_{\Wb}(\xb)} \\
  & = 2\EE_{\hat\cD}\sbr{\ell'(f^Q_{\Wb}(\xb), y)
    \cdot (f^Q_{\Wb}(\xb) - f^Q_{\boldsymbol{0}}(\xb))} \\
& \stackrel{(i)}{\ge}
    2(\hat\cR(f^Q_\Wb) - \hat\cR(f^Q_{\boldsymbol{0}}))\\& \stackrel{(ii)}{\ge} -2,
\end{align*}
where (i) uses convexity of $\ell$ and (ii) uses the assumption that
$\ell(0, y)\le 1$ for all $y\in\cY$. 

Combining with the fact that $\inner{\nabla_\Wb (\norm{\Wb}_{2, 4}^4)}{\Wb}=4\norm{\Wb}_{2, 4}^4$, we have simultaneously for all $\Wb$ that
\begin{align*}
\inner{\nabla \hat{\cR}_{\lambda}(f^Q_\Wb)}{\Wb} & \ge \inner{\nabla_\Wb (\lambda\norm{\Wb}_{2, 4}^4)}{\Wb} + \inner{\nabla_\Wb \hat{\cR}(f_\Wb^Q)}{\Wb} \\
& \ge 4\lambda\norm{\Wb}_{2, 4}^4 -2.
\end{align*}
Therefore we see that any stationary point $\Wb$ has to satisfy
\begin{align*}
\norm{\Wb}_{2, 4} \le (2\lambda)^{-1/4}.
\end{align*}
%
Choosing
\begin{equation*}
  \lambda_0 =\frac{1}{36} (2\tau M + \epsilon),
\end{equation*}
we get
$36 \lambda B_{w, \star}^4=2\tau M + \epsilon$. The Hessian of $\hat{\cR}_\lambda(f^Q_{\Wb})$ along direction $\Wb_\star$ is
\begin{align*}
\nabla^2_{\Wb} \hat{\cR}_\lambda(f^Q_{\Wb})[\Wb_\star, \Wb_\star] & = \nabla^2_{\Wb} \hat{\cR}(f^Q_{\Wb})[\Wb_\star, \Wb_\star] + \lambda \nabla_{\Wb}^2 \norm{\Wb}_{2, 4}^4 [\Wb_\star, \Wb_\star] \\
& = \nabla^2_{\Wb} \hat{\cR}(f^Q_{\Wb})[\Wb_\star, \Wb_\star] + 4\lambda \sum_{r \leq m}\norm{\wb_r}_2^2 \norm{\wb_{\star, r}}_2^2 + 2\inner{\wb_r}{\wb_{\star, r}}^2 \\
& \leq \inner{\nabla \hat\cR(f^Q_\Wb)}{\Wb} - 2(\hat\cR(f^Q_\Wb) - M) + m^{-1} C^2B_h^4\norm{\Wb}_{2, 4}^2 \norm{\Wb_\star}_{2, 4}^2 \\
& \quad + 12 \lambda \norm{\Wb}_{2, 4}^2 \norm{\Wb_\star}_{2, 4}^2 \\
& \overset{(i)}{\leq} \inner{\nabla \hat\cR(f^Q_\Wb)}{\Wb} - 2(\hat\cR(f^Q_\Wb) - M) + m^{-1} C^2B_h^4\norm{\Wb}_{2, 4}^2 \norm{\Wb_\star}_{2, 4}^2 \\
& \quad + \lambda \norm{\Wb}_{2, 4}^4 + 36\lambda\norm{\Wb_\star}_{2, 4}^4 \\
& \leq \inner{\nabla \hat\cR_\lambda(f^Q_\Wb)}{\Wb} - 2(\hat\cR_\lambda(f^Q_\Wb) - M) + m^{-1} C^2B_h^4\norm{\Wb}_{2, 4}^2 \norm{\Wb_\star}_{2, 4}^2 \\
& \quad - \lambda \norm{\Wb}_{2, 4}^4 + 36\lambda \norm{\Wb_\star}_{2, 4}^4. 
\end{align*}
We used the fact $12 ab \leq a^2 + 36b^2$. For a second order-stationary point $\hat\Wb$ of $\hat\cR_\lambda(f^Q_\Wb)$, its gradient vanishes and the Hessian is possitive definite. Therefore, we have
\begin{align*}
0 \leq - 2(\hat\cR_\lambda(f^Q_{\hat\Wb}) - M) + m^{-1} C^2B_h^4\norm{\hat\Wb}_{2, 4}^2 \norm{\Wb_\star}_{2, 4}^2 - \lambda \norm{\hat\Wb}_{2, 4}^4 + 36\lambda \norm{\Wb_\star}_{2, 4}^4. 
\end{align*}
We choose $m = \epsilon^{-1}(2\lambda_0)^{-1/2} C^2 B_h^4 B_{w, \star}^4 \geq \epsilon^{-1}C^2 B_h^4\|\hat\Wb\|_{2, 4}^2 \norm{\Wb_\star}_{2, 4}^2$ and the above inequality implies
\begin{align*}
& 2(\hat\cR_\lambda(f^Q_{\hat\Wb}) - M) \leq 2\tau M + \epsilon + \epsilon \\
\Longrightarrow ~& \hat\cR_\lambda(f^Q_{\hat\Wb}) \leq (1+\tau)M + \epsilon.
\end{align*}
The proof is complete.
\end{proof}

\subsection{Proof of Generalization in Theorem \ref{thm:opt-gen}}
\begin{proof}[Proof of Theorem \ref{thm:opt-gen}, Generalization Part]\label{pf:gen}
Using symmetrization, we have
\begin{align*}
\EE_{(\xb_i, y_i)} \sbr{\sup_{\norm{\Wb}_{2, 4} \le B_w} \left|\cR(f^Q_{\Wb}) - \hat{\cR}(f^Q_{\Wb})\right|} \le 2\EE_{(\xb_i, y_i),\boldsymbol{\xi}} \sbr{\sup_{\norm{\Wb}_{2,4} \le B_w} \left|\frac{1}{n} \sum_{i=1}^n \xi_i \ell(f^Q_\Wb(\xb_i), y_i)\right|},
\end{align*}
where $\xi$ is i.i.d. Rademacher random variables. The above Rademacher complexity can be bounded using the contraction theorem~\citep[Chapter 5]{wainwright2019high}:
\begin{align*}
&\quad~  \EE_{(\xb_i, y_i), \boldsymbol{\xi}} \sbr{\left|\sup_{\norm{\Wb}_{2, 4} \le B_w} \frac{1}{n} \sum_{i=1}^n \xi_i \ell(y_i, f^Q_\Wb(\xb_i))\right|} \\
& = \EE_{(\xb_i, y_i), \boldsymbol{\xi}} \sbr{\sup_{\norm{\Wb}_{2, 4} \le B_w} \max\set{\frac{1}{n} \sum_{i=1}^n \xi_i \ell(y_i, f^Q_\Wb(\xb_i)), - \frac{1}{n} \sum_{i=1}^n \xi_i \ell(y_i, f^Q_\Wb(\xb_i))}} \\
& \leq \EE_{(\xb_i, y_i), \boldsymbol{\xi}} \sbr{\sup_{\norm{\Wb}_{2, 4} \le B_w} \frac{1}{n} \sum_{i=1}^n \xi_i \ell(y_i, f^Q_\Wb(\xb_i)) + \sup_{\norm{\Wb}_{2, 4} \leq B_w} \frac{1}{n} \sum_{i=1}^n - \xi_i \ell(y_i, f^Q_\Wb(\xb_i))} \\
& \le 4\EE_{(\xb_i, y_i), \boldsymbol{\xi}}\sbr{\sup_{\norm{\Wb}_{2, 4}\le B_w} \frac{1}{n}\sum_{i=1}^n \xi_i f^Q_\Wb(\xb_i)} + 2\EE_{(\xb_i, y_i), \boldsymbol{\xi}}\sbr{\frac{1}{n}\sum_{i=1}^n \xi_i\ell(0, y_i)} \\
& \le 4\EE_{\xb_i, \boldsymbol{\xi}}\sbr{\sup_{\norm{\Wb}_{2, 4}\le B_w} \frac{1}{\sqrt{m}}\sum_{r\le m} 
    \inner{\frac{1}{n} \sum_{i=1}^n\xi_i a_{r}\phi''(\wb_{0,r}^\top\hb(\xb_i))\hb(\xb_i)\hb(\xb_i)^\top}{\wb_r\wb_r^\top}} + \frac{2}{\sqrt{n}} \\
& \le 4\EE_{\xb_i, \boldsymbol{\xi}}\sbr{\sup_{\norm{\Wb}_{2, 4}\le B_w}
    \max_{r\in[m]}\opnorm{\frac{1}{n}\sum_{i=1}^n
    \xi_i\phi''(\wb_{0,r}^\top\hb(\xb_i))\hb(\xb_i)\hb(\xb_i)^\top} \cdot
    \frac{1}{\sqrt{m}}\sum_{r\le m}\norm{\wb_r\wb_r^\top}_{*}} +
    \frac{2}{\sqrt{n}} \\
& \le 4\EE_{\xb_i, \boldsymbol{\xi}}\sbr{\max_{r\in[m]}\opnorm{\frac{1}{n}\sum_{i=1}^n
    \xi_i\phi''(\wb_{0,r}^\top \hb(\xb)_i) \hb(\xb_i)\hb(\xb_i^\top)}} \cdot
    \underbrace{\sup_{\norm{\Wb}_{2, 4}\le B_w} \frac{1}{\sqrt{m}}\sum_{r\le m}\norm{\wb_r}_2^2}_{\le B_w^2} +
    \frac{2}{\sqrt{n}},
\end{align*}
where the last step used the power mean (or Cauchy-Schwarz)
inequality on $\set{\norm{\wb_r}_2}$ and $\norm{\cdot}_{*}$ denotes the matrix nuclear norm (sum of singular values). Now it only remains to bound the expected max operator norm above. We apply the matrix concentration lemma~\citet[Lemma 8]{bai2019beyond} to deduce that
\begin{align*}
& \EE_{\xb_i, \boldsymbol{\xi}}\sbr{ \max_{r\in[m]}
    \opnorm{\frac{1}{n}\sum_{i=1}^n 
    \xi_i\phi''(\wb_{0,r}^\top\hb(\xb_i))\hb(\xb_i)\hb(\xb_i)^\top} } \\
& \le 4\sqrt{\log(2Dm)} \cdot
  \EE_{\xb_i}\sbr{\sqrt{\max_{r\in[m]}
  \opnorm{\frac{1}{n^2}\sum_{i=1}^n
    \phi''(\wb_{0,r}^\top\hb(\xb_i))^2\norm{\xb_i}_2^2 \hb(\xb_i)\hb(\xb_i)^\top}}} \\
  & \le 4B_h\sqrt{\frac{\log(2Dm)}{n}} \cdot \EE_{\xb_i}\sbr{
    \sqrt{\max_{r,i}\phi''(\wb_{0,r}^\top\hb(\xb_i))^2  \cdot
    \opnorm{\frac{1}{n}\sum_{i=1}^n \hb(\xb_i)\hb(\xb_i)^\top}}} \\
  & \le 4B_h\sqrt{\frac{\log(2Dm)}{n}}
    \rbr{\EE_{\xb_i}\sbr{\max_{r,i}\phi''(\wb_{0,r}^\top\hb(\xb_i))^2}
    \cdot
    \EE_{\xb_i}\sbr{\opnorm{\frac{1}{n}\sum_{i=1}^n\hb(\xb_i)\hb(\xb_i)^\top}}}^{1/2} \\
& \leq 4C^2B_h\sqrt{\frac{\log(2Dm)}{n}}
    \EE_{\xb_i}\sbr{\opnorm{\frac{1}{n}\sum_{i=1}^n\hb(\xb_i)\hb(\xb_i)^\top}}^{1/2}.
\end{align*}
Combining all the ingredients and substituting $M_{h, \textrm{op}} = B_h^{-1} \EE_{\xb_i}\sbr{\opnorm{\frac{1}{n}\sum_{i=1}^n\hb(\xb_i)\hb(\xb_i)^\top}}^{1/2}$, the generalization error is bounded by
\begin{align*}
\EE_{(\xb_i, y_i)}\sbr{\sup_{\norm{\Wb}_{2, 4} \leq B_w} \left|\cR(f^Q_{\Wb}) - \hat{\cR}(f^Q_{\Wb}) \right|} \leq \tilde{O}\left(\frac{B_h^2 B_w^2 M_{h, \textrm{op}}}{\sqrt{n}} \sqrt{\log (Dm)}+ \frac{1}{\sqrt{n}}\right).
\end{align*}
\end{proof}

\section{Results on feature covariance}
\label{appendix:covariance}

\subsection{Technical tool}
\label{appendix:covariance-tool}

We first present a Lemma for relating the covariance of nonlinear random features to the covariance of certain polynomial bases, adapted from~\citep[Proposition 2]{ghorbani2019linearized}.
\begin{lemma}[Covariance through polynomials]
  \label{lemma:covariance-polynomials}
  Let $\vb_i\simiid{\rm Unif}(\SSS^{d-1})$ be random unit vectors for
  $i\in[D]$ and $\Vb=[\vb_1,\dots,\vb_D]^\top\in\R^{D\times d}$.
  \begin{enumerate}[wide, label=(\alph*)]
  \item For any $k\ge 0$, suppose $D\le O(d^{k+1-\delta})$ for some
    $\delta>0$, then we have with
    high probability as $d\to\infty$ that
    \begin{align*}
      \lambda_{\min}\paren{ (\Vb\Vb^\top)^{\odot (k+1)} } \ge \frac{1}{2},
    \end{align*}
    where $(\cdot)^{\odot k}$ is the Hadamard product: $(\Ab^{\odot k})_{ij} =
    \Ab_{ij}^k$.
  \item In the same setting as above, let $\xb\sim{\rm
      Unif}(\SSS^{d-1})$ and define $\bSigma\in\R^{D\times D}$ with
    \begin{align*}
      \bSigma_{ij} = \E_{\xb} \brac{ \indic{\vb_i^\top\xb\ge 0}
      \indic{\vb_j^\top\xb\ge 0} },
    \end{align*}
    then we have
    \begin{align*}
      \lambda_{\min}(\bSigma) \ge \frac{1}{2} \norm{{\sf P}_{\ge
      k+1}\sigma_d}_{L_2}^2,
    \end{align*}
    where $\sigma_d:{\rm Unif}(\SSS^{d-1}(\sqrt{d}))\to\R$ is defined
    as $\sigma_d(\xb)\defeq \indic{x_1\ge 0}$, and ${\sf P}_{\ge k+1}$ 
    denotes the projection onto degree $\ge (k+1)$ polynomials under
    the base measure ${\rm Unif}(\SSS^{d-1}(\sqrt{d}))$.
  \end{enumerate}
\end{lemma}

\subsection{Lower bound on population covariance}
\label{appendix:population-covariance}
We first present a lower bound when $\xb\sim{\rm Unif}(\SSS^{d-1})$ is
uniform on the sphere, and when the features are the biasless
indicator random features.
\begin{lemma}[Lower bound of population covariance]\label{lemma:covariance}
  \label{lemma:covariance-lower-bound}
  Let $\xb\sim {\rm Unif}(\mathbb{S}^{d-1})$ and suppose we sample
  $\vb_i\simiid \normal(\boldsymbol{0}, \Ib_d)$ for $1\le i\le D$, where $D\le
  O(d^K)$. Let $\bSigma\in\R^{D\times D}$ be the (population)
  covariance matrix of the random features $\set{\indic{\vb_i^\top
      \xb\ge 0}}_{i\in[D]}$, that is,
  \begin{align*}
    \bSigma_{ij} \defeq \E_{\xb\sim {\rm Unif}(\mathbb{S}^{d-1})}\brac{
    \indic{\vb_i^\top \xb\ge 0 } \indic{\vb_j^\top \xb\ge 0}},
  \end{align*}
  then we have $\lambda_{\min}(\bSigma)\ge c>0$ with high
  probability as $d\to\infty$, where $c=c_K$ is a constant that
  depends on $K$ (and the indicator activation) but not $d$.
\end{lemma}
\begin{proof}
  Let $\wt{\vb}_i\defeq \vb_i/\norm{\vb_i}_2$ denote the normalized
  version of $\vb_i$, then $\wt{\vb}_i\simiid {\rm
    Unif}(\mathbb{S}^{d-1})$ due to the spherical symmetry of
  $\normal(\boldsymbol{0}, \Ib_d)$. Further using the positive homogeneity of
  $t\mapsto \indic{t\ge 0}$ yields that
  \begin{align*}
    \bSigma_{ij}& = \E_{\xb\sim {\rm Unif}(\mathbb{S}^{d-1})}
    \brac{\indic{\wt{\vb}_i^\top\xb \ge 0} \indic{\wt{\vb}_j^\top \xb \ge
    0}} \\&= \E_{\xb\sim {\rm Unif}(\mathbb{S}^{d-1}(\sqrt{d}))}
    \brac{\indic{\wt{\vb}_i^\top\xb \ge 0} \indic{\wt{\vb}_j^\top \xb \ge 0}}.
  \end{align*}
  This falls into the setting of
  Lemma~\ref{lemma:covariance-polynomials}(b), applying which implies
  that with high probability (as $d\to\infty$) we have
  \begin{align*}
    \lambda_{\min}(\bSigma) \ge \frac{1}{2} \norm{ {\sf P}_{\ge
    K+1}\sigma_d}_{L_2}^2,
  \end{align*}
  where $\sigma_d:\mathbb{S}^{d-1}(\sqrt{d})\to\R$ is defined as
  $\sigma_d(\xb) = \indic{x_1\ge 0}$, ${\sf P}_{\ge K+1}$ denotes the
  projection onto degree-$\ge K+1$ polynomials under the base measure
  ${\rm Unif}(\mathbb{S}^{d-1}(\sqrt{d}))$, and the $L_2$ norm is under
  the same base measure. For large enough $d$, as
  $x_1|\xb\sim{\rm Unif}(\mathbb{S}^{d-1}(\sqrt{d})) \Rightarrow
  \normal(0,1)\defeq \gamma$ (where $\Rightarrow$ denotes convergence
  in distribution), this is further lower bounded by
  \begin{align}
    \label{equation:spherical-lower-bound}
    \frac{1}{4} \norm{{\sf P}_{\ge K+1}\indic{\cdot \ge 0}}_{L_2(\gamma)}^2 = c_K > 0
  \end{align}
  as the indicator function is not a polynomial of any degree (so
  that its $L_2$ projection onto polynomials of degree $\le K$ is not
  itself for any $K\ge 0$).

  \textbf{Decay of eigenvalue lower bounds with uniform data}.

  We now provide a lower bound for the quantity
  $\norm{{\sf P}_{\ge K+1}\indic{\cdot \ge
      0}}_{L_2(\gamma)}^2$, thereby giving a lower bound on $c_K$
  defined in~\eqref{equation:spherical-lower-bound}. Indeed, we have
  \begin{align*}
    \norm{{\sf P}_{\ge K+1}\indic{\cdot \ge
      0}}_{L_2(\gamma)}^2 = \sum_{j=K+1}^{\infty} \hat{\sigma}_j^2,
  \end{align*}
  where
  \begin{align*}
    \indic{z\ge 0} \stackrel{L_2(\gamma)}{=} \sum_{j=0}^\infty
    \hat{\sigma}_j h_j(z)
  \end{align*}
  is the Hermite decomposition of $\indic{z\ge 0}$.
  By~\citep{kalai2008agnostically}, we know that
  \begin{align*}
    \hat{\sigma}_0 = \frac{1}{2}, \quad \hat{\sigma}_{2i+1} = (-1)^i
    \sqrt{\frac{1}{2\pi(2i+1)!}} \frac{(2i)!}{2^i i!}, \quad
    \hat{\sigma}_{2i+2}  = 0~~\textrm{for all}~i\ge 0.
  \end{align*}
  We now calculate the decay of $\hat{\sigma}_{2i+1}^2$. By Stirling's
  formula, we have
  \begin{align*}
    \hat{\sigma}_{2i+1}^2 = \frac{1}{2\pi(2i+1)!} \cdot
    \frac{(2i)!^2}{2^{2i}(i!)^2} \asymp \frac{1}{2\pi (2i+1)} \cdot
    \frac{\sqrt{2\pi\cdot 2i}(2i/e)^{2i}}{2^{2i} \cdot 2\pi i \cdot
    (i/e)^{2i}} \asymp C i^{-3/2}
  \end{align*}
  for some absolute constant $C>0$. This means that for all $i\ge 0$
  we have $\hat{\sigma}_{2i+1}^2 \ge Ci^{-3/2}$ for some (other)
  absolute constant $C>0$, which gives
  \begin{align*}
    \sum_{j=K+1}^\infty \hat{\sigma}_j^2 \ge \sum_{i:2i+1\ge
    K+1}^\infty Ci^{-3/2} \ge CK^{-1/2}.
  \end{align*}
  Therefore we have $c_K\ge \Omega(K^{-1/2})$ for all $K$.
\end{proof}

{\bf Covariance lower bounds for non-uniform data}.

We further show that Assumption \ref{assumption_covariance} can hold fairly generally when $\xb$ is no longer uniform on $\SSS^{d-1}$. Recall we choose $D \leq O(d^K)$ for some constant $K$. 

We begin with assuming there exists a positive definite matrix $\Sbb \in \RR^{d \times d}$ such that $\xb$ is equal in distribution as $\Sbb^{1/2}\zb / \norm{\Sbb^{1/2}\zb}_2$, where $\zb \sim \normal(\boldsymbol{0}, \Ib_d)$. In other words, $\xb$ is distributed as a rescaled version of a $d$-dimensional Gaussian with \emph{arbitrary covariance}, a fairly expressive set of distributions which can model the case where $\xb$ is far from uniform over the sphere. We show in this case that Assumption~\ref{assumption_covariance} holds, whenever $\Sbb$ has a bounded condition number (i.e. $\lambda_{\min}(\Sbb) / \lambda_{\max}(\Sbb)\ge 1/\kappa$ where $\kappa>0$ does not depend on $d$).

Indeed, we can deduce
\begin{align*}
\ind\{\vb_j^\top \xb \geq 0\} = \ind\set{\vb_j^\top \frac{\Sbb^{1/2}\zb}{\norm{\Sbb^{1/2}\zb}_2} \geq 0} = \ind\set{(\Sbb^{1/2}\vb_j)^\top \zb \geq 0} = \ind\set{\frac{\Sbb^{1/2}\vb_j^\top}{\norm{\Sbb^{1/2}\vb_j}_2} \zb \geq 0}.
\end{align*}
Here the equality denotes two random variables following the same distribution. We apply the Hermite decomposition of indicator function to decompose the covariance matrix $\bSigma$:
\begin{align}\label{eq:sigmadecompose}
\lambda_{\min}(\bSigma) & = \min_{\norm{\ub}_2 = 1} \EE_{\xb} \left[\ub^\top\gb(\xb) \gb(\xb)^\top\ub\right] \notag\\
& = \min_{\norm{\ub}_2 = 1} \EE_{\xb} \left[\sum_{i, j} \ind\{\vb_i^\top \xb \geq 0\} \ind\{\vb_j^\top \xb \geq 0\} u_i u_j \right] \notag \\
& = \min_{\norm{\ub}_2 = 1} \sum_{i, j} \left(\cT_1 + \cT_2\right)u_i u_j,
\end{align}
where $\cT_1$ and $\cT_2$ are given as follows,
\begin{align*}
& \cT_1 = \sum_{\ell=0}^\infty \hat\sigma_\ell^2 \EE_\zb\left[h_\ell\left(\frac{\Sbb^{1/2}\vb_j^\top}{\norm{\Sbb^{1/2}\vb_j}_2} \zb\right)h_\ell\left(\frac{\Sbb^{1/2}\vb_j^\top}{\norm{\Sbb^{1/2}\vb_j}_2} \zb\right)\right]\\& = \sum_{\ell=0}^\infty \hat\sigma_\ell^2 \left(\frac{\vb_i^\top\Sbb\vb_j}{\norm{\Sbb^{1/2}\vb_j}_2\norm{\Sbb^{1/2}\vb_i}_2}\right)^\ell, \\
& \cT_2 = \sum_{\ell \neq k} \hat\sigma_\ell \hat\sigma_k \EE_{\zb} [h_\ell(\vb_i^\top \zb) h_k(\vb_j^\top \zb)] = 0,
\end{align*}
where $\hat\sigma_\ell$ is the coefficient of Hermite decomposition of the indicator function, and $\cT_2$ vanishes, due to the orthogonality of probabilistic Hermite polynomials. We proceed to bound the minimum singular value of $\bSigma$:
\begin{align*}
\lambda_{\min}(\bSigma) & = \min_{\norm{\ub}_2 = 1} \sum_{i, j} \sum_{\ell=0}^\infty \hat\sigma_\ell^2 \left(\frac{\vb_i^\top\Sbb\vb_j}{\norm{\Sbb^{1/2}\vb_j}_2\norm{\Sbb^{1/2}\vb_i}_2}\right)^\ell u_i u_j. 
\end{align*} 
Note that in the above decomposition, $\bSigma$ is the sum of an infinite series of positive semidefinite matrices. To show $\bSigma$ has a lower bounded smallest singular value, it suffices to show that there exists a summand in the infinite series being positive definite. We confirm this by analyzing the $\ell = K+1$ summand. In fact, we show the following matrix $\bSigma_{K+1} \in \RR^{D \times D}$ is positive definite and its smallest singular value is lower bounded by some constant independent of $d$.
\begin{align*}
[\bSigma_{K+1}]_{ij} = \hat\sigma_{K+1}^2 \left(\frac{\vb_i^\top\Sbb\vb_j}{\norm{\Sbb^{1/2}\vb_j}_2\norm{\Sbb^{1/2}\vb_i}_2}\right)^{K+1} \quad \textrm{for}\quad i, j = 1, \dots, D.
\end{align*}
We denote the normalized $\vb_j$ as $\tilde\vb_j = \vb_j / \norm{\vb_j}_2$, and derive a lower bound on the singular value of $\tilde\bSigma_{K+1}$ with
\begin{align*}
[\tilde\bSigma_{K+1}]_{ij} = (\tilde\vb_i^\top \Sbb \tilde\vb_j)^{K+1}.
\end{align*}
Using the tensor product notation, we rewrite $\tilde\bSigma_{K+1}$ as
\begin{align*}
\tilde\bSigma_{K+1} = \tilde\Vb^{\ast (K+1)} \Sbb^{\otimes (K+1)} \left(\tilde\Vb^{\ast (K+1)}\right)^\top,
\end{align*}
where $\tilde\Vb = [\tilde\vb_1, \dots, \tilde\vb_D]^\top \in \RR^{D \times d}$, $\tilde\Vb^{\ast (K+1)} \in \RR^{D \times d^{K+1}}$ is the Khatri-Rao product, and $\Sbb^{\otimes (K+1)} \in \RR^{d^{K+1} \times d^{K+1}}$ denotes the Kronecker product. Then we know
\begin{align*}
\tilde\bSigma_{K+1} \succeq \lambda_{\min}^{K+1}(\Sbb) \tilde\Vb^{\ast (K+1)} \left(\tilde\Vb^{\ast (K+1)}\right)^\top.
\end{align*}
Moreover, using Lemma~\ref{lemma:covariance-polynomials}(a), we have
$\lambda_{\min}\left(\tilde\Vb^{\ast (K+1)}
  \left(\tilde\Vb^{\ast (K+1)}\right)^\top\right) \geq 1/2$ as we picked $D\le O(d^K)$. Substituting into $\bSigma_{K+1}$, we have 
\begin{align*}                                                                                                                                        \lambda_{\min}(\bSigma_{K+1}) & \geq \frac{1}{2} \hat\sigma_{K+1}^2 \lambda_{\min}^{K+1}(\Sbb) \left(\frac{\norm{\vb_i}_2 \norm{\vb_j}_2}{\norm{\Sbb^{1/2}\vb_i}_2 \norm{\Sbb^{1/2}\vb_j}_2}\right)^{K+1} \\
                                                                                                                                                                      & \geq \frac{1}{2}\hat\sigma_{K+1}^2 \underbrace{\lambda_{\min}^{K+1}(\Sbb) \lambda_{\max}^{-(K+1)}(\Sbb)}_{\kappa^{-(K+1)}}.  \end{align*}
Therefore the smallest singular value of $\bSigma$ is lower bounded by $\Omega(\hat\sigma_{K+1}^2 \kappa^{-(K+1)})$, which is a constant only depending on $K$ but not $d$. This finishes the proof. \qed 

We remark that Assumption~\ref{assumption_covariance} can hold much more generally than rescaled Gaussian distributions, provided the following set of conditions holds: There exists some random variable $\zb\in\R^d$ such that $\xb$ is equal in distribution to $\zb/\ltwo{\zb}$, a set of univariate polynomials $\{h^{(i)}_k:\R\to\R\}$ for $i = 1, \dots, D$ and $k\ge 0$ where each $h^{(i)}_k$ is a degree-$k$ polynomial ``assigned'' to $\vb_i$, and corresponding coefficients $\hat{\sigma}^{(i)}_k\in\R$, such that
\begin{align*}
  \indic{\vb_i^\top \zb\ge 0} \stackrel{L_2}{=} \sum_{k=0}^\infty \hat{\sigma}_k^{(i)} h^{(i)}_k(\vb_i^\top \zb).
\end{align*}
Let $\bSigma_k,\bSigma_{k,>k},\bSigma_{>k}\in\R^{D\times D}$ be defined as
\begin{align*}
  & [\bSigma_k]_{ij} \defeq \E_{\zb}\brac{\hat{\sigma}_k^{(i)}\hat{\sigma}_k^{(j)} h^{(i)}_{k}(\tilde{\vb}_i^\top\zb) h^{(j)}_{k}(\tilde{\vb}_j^\top\zb)} \\
  & [\bSigma_{k,>k}]_{ij} \defeq \E_{\zb} \brac{\hat{\sigma}_k^{(i)}h^{(i)}_k(\tilde{\vb}_i^\top \zb) \cdot \sum_{\ell>k} \hat{\sigma}^{(j)}_\ell h^{(j)}_\ell(\tilde{\vb}_j^\top \zb)} \\
  & [\bSigma_{>k}]_{ij} \defeq \E_{\zb} \brac{ \sum_{\ell>k} \hat{\sigma}^{(i)}_\ell h^{(i)}_\ell(\tilde{\vb}_i^\top \zb) \cdot \sum_{\ell>k} \hat{\sigma}^{(j)}_\ell h^{(j)}_\ell(\tilde{\vb}_j^\top \zb) },
\end{align*}
where $\tilde{\vb}_i\defeq\vb_i/\ltwo{\vb_i}$. We assume we have
\begin{itemize}[wide]
\item The not-too-correlated condition: there exists some $\epsilon\in(0,1]$ such that for any $k\ge 0$, we have
  \begin{align*}
    \bSigma_{\ge k} \defeq \bSigma_k + \bSigma_{k, >k} + \bSigma_{k,
    >k}^\top + \bSigma_{>k} \succeq \epsilon(\bSigma_k + \bSigma_{>k}).
  \end{align*}
\item For all large $d$ and $D\le O(d^k)$, we have
  \begin{align*}
    \bSigma_{k+1} \succeq c_{k+1} (\tilde{\Vb}\tilde{\Vb}^\top)^{\odot (k+1)},
  \end{align*}
  where $c_{k+1}>0$ is a constant that depends on $k$ but not $d$.
\end{itemize}
In this case, we can deduce that for $D = O(d^K)$, Assumption~\ref{assumption_covariance} holds with $\lambda_K = \Omega(\epsilon^{K+2}c_{K+1})$. To see this, observe that we have the expansion $\indic{t\ge 0} = \hat{\sigma}^{(i)}_0h^{(i)}_0(t) + \sum_{\ell>0}\hat{\sigma}^{(i)}_\ell h^{(i)}_\ell(t)$ for all $i=1, \dots, D$, and thus
\begin{align*}
  \bSigma = \bSigma_{0} + \bSigma_{0, >0} + \bSigma_{0, >0}^\top +
  \bSigma_{>0, >0} \stackrel{(i)}{\succeq} \epsilon(\bSigma_{0} + \bSigma_{>0}) \succeq \epsilon\bSigma_{>0},
\end{align*}
where $(i)$ applied the not-too-correlated condition. Repeating the above process for $K$ times leads to
\begin{align*}
  \bSigma \succeq \epsilon^{K+1} \bSigma_{>K} \succeq \epsilon^{K+2}
  (\bSigma_{K+1} + \bSigma_{>(K+1)}) \succeq \epsilon^{K+2}\bSigma_{K+1}
  \succeq \epsilon^{K+2} c_{K+1} (\tilde{\Vb} \tilde{\Vb}^\top)^{\odot (K+1)}.
\end{align*}
Combining with existing lower bound $\lambda_{\min}((\tilde{\Vb}\tilde{\Vb}^\top)^{\odot (K+1)})\ge 1/2$ (Lemma~\ref{lemma:covariance-polynomials}(a)), we see Assumption~\ref{assumption_covariance} holds with $\lambda_K=\Omega(\epsilon^{K+2}c_{K+1})$, a constant that depends on $K$ and independent of $d$. 

We further note that the above two conditions are all satisfied by the rescaled Gaussian distributions: Choosing $h^{(i)}_k\equiv h_k$ (the $k$-th Hermite polynomial) for all $i = 1, \dots, D$, the first condition holds with $\epsilon=1$ since $\bSigma_{k,>k}=\zero$, and the second condition holds with $c_{k+1} = (\lambda_{\min}(\Sbb)/\lambda_{\max}(\Sbb))^{k+1}$ (as shown earlier). Combining with the fact they only assume things about the moments of $\xb$ (or $\zb$; since $h_k^{(i)}$ are polynomials), we see that they are indeed moment-based assumptions that contain Gaussian distributions with arbitrary covariances, and thus can be fairly general.

We also remark that while we have verified Assumption~\ref{assumption_covariance} for random features without biases, our analyses can be straightforwardly generalized to the case with bias by looking at the augmented input $[\xb^\top, 1]^\top\in\R^{d+1}$ and analyzing its distributions in similar fashions.



\subsection{Relative concentration of covariance estimator}
\label{appendix:covariance-concentration}
\begin{lemma}[Relative concentration of covariance estimator]\label{lemma:covariance_relative}
  Let $\set{\gb(\xb_i)\in\R^D}_{i = 1}^n$ be i.i.d. random vectors such that $\ltwo{\gb_1}\le B_g$ almost surely and $\E[\gb_1\gb_1^\top]=\bSigma\succeq \lambda_{\min}\Ib_D$. Let $\what{\bSigma} \defeq \frac{1}{n}\sum_{i=1}^n \gb(\xb_i)\gb(\xb_i)^\top$ denote the empirical covariance matrix of $\set{\gb(\xb_i)}$. For any $\epsilon\in(0,1)$, as soon as $n\ge C\epsilon^{-2}\lambda_{\min}^{-1}B_g^2\log(n\vee D)$, we have
  \begin{align*}
    \E \brac{ \opnorm{\bSigma^{-1/2}\what{\bSigma}\bSigma^{-1/2} - \Ib_D} } \le \epsilon.
  \end{align*}
  Further, when $n\ge C\delta^{-2}\epsilon^{-2}\lambda_{\min}^{-1}B_g^2\log (n\vee D)$ we have with probability at least $1-\delta$ that
  \begin{align*}
    \opnorm{ \bSigma^{-1/2}\what{\bSigma}\bSigma^{-1/2} - \Ib_D } \le
    \epsilon, 
  \end{align*}
  where $C>0$ is a universal constant. On the same event, we have the relative concentration
  \begin{align*}
    (1-\epsilon)\bSigma \preceq \what{\bSigma} \preceq (1+\epsilon)\bSigma.
  \end{align*}
\end{lemma}
\begin{proof}
  The first statement directly yields the second by the Markov inequality.
  To see how the second statement implies the third, we can left- and
  right- multiply the matrix inside by $\bSigma^{1/2}\vb$ for any
  $\vb\in\R^D$ and get that
  \begin{align*}
  \abs{ (\vb^\top\bSigma^{1/2})\bSigma^{-1/2}\what{\bSigma}\bSigma^{-1/2}(\bSigma^{1/2}\vb) - \vb^\top\bSigma\vb} = \abs{\vb^\top\what{\bSigma}\vb -\vb^\top\bSigma\vb} \le \epsilon \vb^\top \bSigma\vb,
  \end{align*}
  which implies that $(1-\epsilon)\bSigma \preceq \what{\bSigma} \le (1+\epsilon)\bSigma$.
  
  We now prove the first statement, which builds on the following Rudelson's inequality for controlling expected deviation of heavy-tailed sample covariance matrices:
  \begin{lemma}[Restatement of Theorem
    5.45,~\citep{vershynin2010introduction}]
    \label{lemma:rudelson}
    Let $\set{\ab_i\in\R^D}_{i=1}^n$ be independent random vectors with 
    $\E[\ab_i\ab_i^\top]=\Ib_D$. Let $\Gamma\defeq \E[\max_{i\in[n]}
    \ltwo{\ab_i}^2]$. Then there exists a universal constant $C>0$ such
    that letting $\delta\defeq C\Gamma\log(n\vee D)/n$, we have
    \begin{align*}
      \E\brac{ \opnorm{\frac{1}{n}\sum_{i=1}^n \ab_i\ab_i^\top - \Ib_D}}
      \le \delta \vee \sqrt{\delta}.
    \end{align*}
  \end{lemma}

  We will apply Lemma~\ref{lemma:rudelson} on the whitened random vectors $\hb(\xb_i)\defeq \bSigma^{-1/2}\gb(\xb_i)$ (Here we slightly abust the notation to denote $\hb(\xb)$ as the whitened feature using the population covariance matrix). Clearly, $\E_{\xb}[\hb(\xb_i)\hb(\xb_i)^\top]=\E_{\xb}[\bSigma^{-1/2}\gb(\xb_i)\gb(\xb_i)^\top\bSigma^{-1/2}]=\bSigma^{-1/2}\bSigma\bSigma^{-1/2}=\Ib_D$. Further, we have 
  \begin{align*}
    \ltwo{\hb(\xb_i)}^2 = \gb(\xb_i)^\top \bSigma^{-1}\gb(\xb_i)^\top \le \lambda_{\min}^{-1} \ltwo{\gb(\xb_i)}^2 \le \lambda_{\min}^{-1}B_g^2
  \end{align*}
  almost surely, and thus $\Gamma\defeq \E_{\xb}[\max_{i\in[n]}\ltwo{\hb(\xb_i)}^2]\le \lambda_{\min}^{-1}B_g^2$. Therefore, $\set{\hb(\xb_i)}$ satisfy the conditions of Lemma~\ref{lemma:rudelson}, from which we obtain
  \begin{align*}
    & \quad \E_{\xb}\brac{ \opnorm{\bSigma^{-1/2}\what{\bSigma}\bSigma^{-1/2} - \Ib_D} } \\
    & = \E_{\xb}\brac{ \opnorm{\frac{1}{n} \sum_{i=1}^n \hb(\xb_i)\hb(\xb_i)^\top - \Ib_D  } } \le \frac{C\lambda_{\min}^{-1}B_g^2\log(n\vee D)}{n} \vee \sqrt{\frac{C\lambda_{\min}^{-1}B_g^2\log(n\vee D)}{n}}.
  \end{align*}
  Therefore, setting $n\ge C\epsilon^{-2}\lambda_{\min}^{-1}B_g^2\log (n\vee D)$, we get that
  \begin{align*}
    \E_{\xb}\brac{ \opnorm{\bSigma^{-1/2}\what{\bSigma}\bSigma^{-1/2} - \Ib_D} }\le \epsilon.
  \end{align*}
  This finishes the proof.  
\end{proof}

\section{Proofs for Section \ref{sec:upperbound}}
\label{pf:upperbound}
This section devotes to the proof of Theorem \ref{thm:quadneural_generalization}. The proof consists of two main parts: expressivity of neural representation (Sections \ref{bottomexpress} and \ref{topexpress}) and generalization property of \quadneural~(Section \ref{quadneuralgen}). Besides, Section \ref{datadependentreg} presents that using data dependent regularizer also achieves improved sample complexity.
\subsection{Expressivity of neural random features}\label{bottomexpress}
\begin{lemma}\label{lemma:bottomexpress}
For a given vector $\boldsymbol{\beta}$ and integer $k \geq 0$, we let $\vb \sim \normal(\boldsymbol{0}, \Ib_d)$ be a standard Gaussian vector and $b \sim \normal(0, 1)$ independent of $\vb$. Then there exists $a(\vb, b)$ such that $\EE_{\vb, b} [a(\vb, b) \mathds{1}\{\vb^\top \xb + b \geq 0\}] = (\boldsymbol{\beta}^\top \xb)^k$ holds for any $\xb \in \SSS^{d-1}$.
\end{lemma}
\begin{proof}
We denote by $H_j(x)$ the $j$-th probabilistic Hermite polynomial. We pick
\begin{align*}
a(\vb, b) = \begin{cases}
c_k H_k(\vb^\top \boldsymbol{\beta}/\norm{\boldsymbol{\beta}}_2) \mathds{1}\{0 < -b < 1/(2k)\}, & \textrm{if $k$ is even} \\
c_k H_k(\vb^\top \boldsymbol{\beta}/\norm{\boldsymbol{\beta}}_2) \mathds{1}\{|b| < 1/(2k)\}, & \textrm{if $k$ is odd}
\end{cases},
\end{align*}
where $c_k$ is a constant to be determined. For a fixed $\xb$, we denote $z_1 = \vb^\top \boldsymbol{\beta}/\norm{\boldsymbol{\beta}}_2$ and $z_2 = \vb^\top \xb$. It is straightforward to check that $z_1, z_2$ is jointly Gaussian with zero mean and $\EE[z_1 z_2] = \boldsymbol{\beta}^\top \xb/\norm{\boldsymbol{\beta}}_2$. We can now deduce that $z_1$ and $(\boldsymbol{\beta}^\top \xb/ \norm{\boldsymbol{\beta}}_2) z_2 + \sqrt{1-(\boldsymbol{\beta}^\top \xb / \norm{\boldsymbol{\beta}}_2)^2} z_3$ follow the same distribution, where $z_3$ is standard Gaussian independent of $z_1$ and $z_2$. For an even $k$, we can check
\begin{align*}
& \EE_{\vb, b}[a(\vb, b) \mathds{1}\{\vb^\top \xb + b \geq 0\}] \\
=~& c_k \EE_{z_1, z_2, b}[H_k(z_1) \mathds{1}\{z_2 + b \geq 0\} \mathds{1}\{0 < -b < 1/(2k)\}] \\
=~& c_k \EE_{z_2, z_3, b}\left[H_k\Big((\boldsymbol{\beta}^\top \xb/ \norm{\boldsymbol{\beta}}_2) z_2 + \sqrt{1-(\boldsymbol{\beta}^\top \xb/ \norm{\boldsymbol{\beta}}_2)^2} z_3\Big) \mathds{1}\{z_2 + b \geq 0\} \mathds{1}\{0 < -b < 1/(2k)\} \right] \\
=~& c_k \EE_b \EE_{z_2, z_3} \left[H_k\Big((\boldsymbol{\beta}^\top \xb / \norm{\boldsymbol{\beta}}_2) z_2 + \sqrt{1-(\boldsymbol{\beta}^\top \xb / \norm{\boldsymbol{\beta}}_2)^2} z_3\Big) \mathds{1}\{z_2 \geq -b\} \mathds{1}\{0 < -b < 1/(2k)\} ~\big|~ b\right] \\
\overset{(i)}{=}~& c_k q_k (\boldsymbol{\beta}^\top \xb)^k \norm{\boldsymbol{\beta}}_2^{-k},
\end{align*}
where $q_k = \EE_b \left[(k-1)!! \frac{\exp(-b^2/2)}{\sqrt{2\pi}} \mathds{1}\{0 < -b < 1/(2k)\} \sum_{j=1, \textrm{~odd}}^{k-1} \frac{(-1)^{(k-1+j)/2}}{j!!} {k/2-1 \choose{(j-1)/2}} b^j \right]$. The equality $(i)$ invokes {\it Lemma $A.6$} in \citet{allen2019learning}. Similarly, for an odd $k$, we have
\begin{align*}
\EE_{\vb, b}[a(\vb, b) \mathds{1}\{\vb^\top \xb + b \geq 0\}] = c_k q_k (\boldsymbol{\beta}^\top \xb)^k \norm{\boldsymbol{\beta}}_2^{-k}
\end{align*}
with $q_k = \EE_b \left[(k-1)!! \frac{\exp(-b^2/2)}{\sqrt{2\pi}} \mathds{1}\{|b| \leq 1/(2k)\} \sum_{j=1, \textrm{~even}}^{k-1} \frac{(-1)^{(k-1+j)/2}}{j!!} {k/2-1 \choose{(j-1)/2}} b^j \right]$. Here we unify the notation to denote $q_k$ as the coefficient for both the even and odd $k$'s. Using {\it Claim C.1} in \citet{allen2019learning}, we can lower bound $p_k$ by $|p_k| \geq \frac{(k-1)!!}{200k^2}$. The proof is complete by choosing $c_k = 1/p_k$, and accordingly, $|c_k| \leq \frac{200k^2}{(k-1)!!} \norm{\boldsymbol{\beta}}_2^k$.
\end{proof}

{\bf From expectation to finite neuron approximation}.
\begin{lemma}\label{lemma:Dbound}
For a given $\epsilon > 0$ and $\delta > 0$, we choose $D =2\times 200^2 k^5 \norm{\boldsymbol{\beta}}_2^{2k} / (\epsilon^2 \delta)$ and independently generate $\vb_{j} \sim \normal(\mathbf{0}, \Ib_d)$ and $b_j \sim \normal(0, 1)$ for $j = 1, \dots, D$. Then with probability at least $1-\delta$, we have
\begin{align*}
\norm{\frac{1}{D} \sum_{j=1}^D a(\vb_j, b_j) \mathds{1}\{\vb_j^\top \xb + b_j \geq 0\} - (\boldsymbol{\beta}^\top \xb)^k}_{L_2} \leq \epsilon.
\end{align*}
\end{lemma}
\begin{proof}
The desired bound can be obtained by Chebyshev's inequality. We bound the second moment of the $L_2$ norm as
\begin{align*}
& \EE_{\vb, b} \norm{\frac{1}{D} \sum_{j=1}^D a(\vb_j, b_j) \mathds{1}\{\vb_j^\top \xb + b_j \geq 0\} - (\boldsymbol{\beta}^\top \xb)^k}_{L_2}^2 \\
= &~ \EE_{\vb, b} \EE_{\xb} \left[\frac{1}{D} \sum_{j=1}^D a(\vb_j, b_j) \mathds{1}\{\vb_j^\top \xb + b_j \geq 0\} - (\boldsymbol{\beta}^\top \xb)^k\right]^2 \\
= &~ \EE_{\xb} \EE_{\vb, b} \left[\frac{1}{D} \sum_{j=1}^D \Big(a(\vb_j, b_j) \mathds{1}\{\vb_j^\top \xb + b_j \geq 0\} - \EE_{\vb_j, b_j} \left[a(\vb_j, b_j) \mathds{1}\{\vb_j^\top \xb + b_j \geq 0\}\right]\Big) \right]^2 \\
= &~ \frac{1}{D^2} \EE_{\xb} \left[\sum_{j=1}^D \EE_{\vb_j, b_j}\Big[a(\vb_j, b_j) \mathds{1}\{\vb_j^\top \xb + b_j \geq 0\} - \EE_{\vb_j, b_j} \left[a(\vb_j, b_j) \mathds{1}\{\vb_j^\top \xb + b_j \geq 0\}\right] \Big]^2\right] \\
= &~ \frac{1}{D} \EE_{\xb, \vb, b} \left[a(\vb, b) \mathds{1}\{\vb^\top \xb + b \geq 0\} - (\boldsymbol{\beta}^\top \xb)^k\right]^2.
\end{align*}
Using Lemma \ref{lemma:bottomexpress}, we have
\begin{align*}
\EE_{\xb, \vb, b} \left[a(\vb, b) \mathds{1}\{\vb^\top \xb + b \geq 0\} - (\boldsymbol{\beta}^\top \xb)^k\right]^2 & = \EE_{\xb, \vb, b} \left[a^2(\vb, b) \mathds{1}\{\vb^\top \xb + b \geq 0\} - \left(\boldsymbol{\beta}^\top \xb\right)^{2k}\right] \\
& \leq \EE_{\xb, \vb, b} \left[c_k^2 H_k^2(\boldsymbol{\beta}^\top \vb / \norm{\boldsymbol{\beta}}_2) + \left(\boldsymbol{\beta}^\top \xb\right)^{2k} \right] \\
& = c_k^2 \sqrt{2\pi} k! + \EE_{\xb}\left[\boldsymbol{\beta}^\top \xb\right]^{2k} \\
& \leq \frac{200^2 k^4}{(k-1)!! (k-1)!!} k! \norm{\boldsymbol{\beta}}_2^{2k} + \norm{\boldsymbol{\beta}}_2^{2k} \\
& \leq 2 \times 200^2 k^5 \norm{\boldsymbol{\beta}}_2^{2k}.
\end{align*}
The last inequality invokes the identity $\frac{k!}{((k-1)!!)^2} + 1 \leq \frac{k!}{(k-1)!} + 1 \leq 2 \frac{k!}{(k-1)!} = 2k$. Therefore, choosing $D = 2\times 200^2 k^5 \norm{\boldsymbol{\beta}}_2^{2k} / (\epsilon^2 \delta)$ gives rise to
\begin{align*}
& \PP\left(\norm{\frac{1}{D} \sum_{j=1}^D a(\vb_j, b_j) \mathds{1}\{\vb_j^\top \xb + b_j \geq 0\} - (\boldsymbol{\beta}^\top \xb)^k}_{L_2} \geq \epsilon \right) \\
\leq~& \epsilon^{-2} \EE_{\xb, b} \left[\norm{\frac{1}{D} \sum_{j=1}^D a(\vb_j, b_j) \mathds{1}\{\vb_j^\top \xb + b_j \geq 0\} - (\boldsymbol{\beta}^\top \xb)^k}_{L_2}^2\right] \\
\leq~& \delta.
\end{align*}
This completes the proof.
\end{proof}

{\bf From single polynomial to sum of polynomials}.
\begin{lemma}\label{lemma:approx_polys}
Given a function $f(\xb) = \sum_{s=1}^{r_{\star}} (\boldsymbol{\beta}_s^\top \xb)^{k_s}$ defined on $\xb \in \SSS^{d-1}$, and positive constants $\epsilon > 0$ and $\delta > 0$, we choose $D \geq \frac{2\times 200^2 r_{\star}^3 \sum_{s=1}^{r_\star}  k_s^5 \norm{\boldsymbol{\beta}_s}_2^{2k_s}}{\epsilon^2 \delta}$, Then there exists scalar $a(\vb_j, b_j)$ for $j = 1, \dots, D$, such that with probability at least $1-\delta$ over independently randomly sampled $\vb_j \sim \normal(\boldsymbol{0}, \Ib_d)$ and $b_j \sim \normal(0, 1)$ for $j = 1, \dots, D$, we have
\begin{align*}
\norm{\frac{1}{D} \sum_{j=1}^{D} a(\vb_j, b_j) \mathds{1}\{\vb_j^\top \xb + b_j \geq 0\} - f(\xb)}_{L_2} \leq \epsilon.
\end{align*}
\end{lemma}
\begin{proof}
We apply Lemma \ref{lemma:bottomexpress} and Lemma \ref{lemma:Dbound} repeatedly for $r_\star$ times. Specifically, for each fixed $s \leq r_{\star}$, Lemma \ref{lemma:bottomexpress} implies that there exists $a_s(\vb, b)$ such that $\EE_{\vb, b}[a_s(\vb, b) \ind\{\vb^\top \xb + b \geq 0\}] = (\boldsymbol{\beta}_s^\top \xb)^{k_s}$. Then we choose $D_s \geq 2\times 200^2 k_s^5 \norm{\boldsymbol{\beta}_s}_2^{2k_s} r_{\star}^3 / (\epsilon^2 \delta)$ so that with probability at least $1-\delta/r_{\star}$, the following $L_2$ bound holds
\begin{align*}
\norm{\frac{1}{D_s} \sum_{j=1}^{D_s} a_s(\vb_j, b_j) \mathds{1}\{\vb_j^\top \xb + b_j \geq 0\} - (\boldsymbol{\beta}_s^\top \xb)^{k_s}}_{L_2} \leq \epsilon / r_{\star}.
\end{align*}
To this end, we set $D = \sum_{s=1}^{r_\star} D_s \geq \frac{2\times 200^2 r_{\star}^3 \sum_{s=1}^{r_\star} k_s^5 \norm{\boldsymbol{\beta}_s}_2^{2k_s}}{\epsilon^2 \delta}$ and define $$\gb(\xb) = [\gb_1(\xb)^\top, \dots, \gb_{r_{\star}}(\xb)^\top]^\top$$ with the $j$-the element of $\gb_s$ as $[\gb_s(\xb)]_j = \ind\{\vb_j^\top \xb + b_j \geq 0\}$ for $j = 1, \dots, D_s$. In other words, we construct a random feature vector $\gb(\xb) \in \RR^D$ by stacking all the random features for approximating the $k_s$-degree polynomial. Similar to $\gb$, we denote $\ab = [\ab_1^\top, \dots, \ab_{r_\star}^\top]^\top$ with the $j$-th element of $\ab_s$ as $[\ab_s]_j = \frac{1}{D_s} a_s(\vb_j, b_j)$ for $j = 1, \dots, D_s$. Then we can bound the $L_2$ distance between $f(\xb)$ and $\ab^\top \gb(\xb)$:
\begin{align*}
\norm{\ab^\top \gb(\xb) - f(\xb)}_{L_2} & \leq \sum_{s=1}^{r_\star} \norm{\ab_s^\top \gb_s(\xb) - (\boldsymbol{\beta}_s^\top \xb)^{k_s}}_{L_2} \\
& \leq \sum_{s=1}^{r_\star} \norm{\frac{1}{D_{s}} \sum_{j=1}^{D_s} a_s(\vb_j, b_j) \ind\{\vb_j^\top \xb + b_j \geq 0\} - (\boldsymbol{\beta}_s^\top \xb)^{k_s}}_{L_2} \\
& \leq \epsilon.
\end{align*}
The above inequality holds with probability $1-\delta$ by the union bound. We complete the proof.
\end{proof}
Lemma \ref{lemma:approx_polys} showcases how to express a sum of polynomials by stacking neural random features for approximating individual polynomials. This technique will be extensively used in the remaining proofs.

\subsection{Expressivity of \quadh}\label{topexpress}
We show \quadneural with neural representation $\hb$ can approximate any function $f$ of the form
\begin{align}
  \label{equation:f-star-appendix}
  f(\xb) = \sum_{s=1}^{r_\star} \alpha_s (\bbeta_s^\top \xb)^{p_s},~~~\textrm{where}~~~|\alpha_s|\le 1,~\norm{(\bbeta_s^\top\xb)^{p_s}}_{L_2}\le 1,~p_s\le p~~\textrm{for all}~s.
\end{align}
To ease the presentation, we temporarily assume all the $p_s$ are even. We extend to odd-degree polynomials in \ref{lemma:odd}. Recall we denote $$\gb(\xb) = [\gb_1(\xb)^\top, \dots, \gb_{r_\star}(\xb)^\top]^\top \quad \textrm{with}~\gb_s(\xb) \textrm{~being a collection of random indicator functions}.$$
We whiten $\gb(\xb)$ by the estimated covariance matrix $\hat{\bSigma}$ to obtain $\hb(\xb) = \hat{\bSigma}^{-1/2}\gb(\xb)$. Note that $\hb(\xb)$ is a $D$-dimensional vector. 
The approximation of \quadh~is stated in the following lemma.
\begin{lemma}\label{lemma:topexpress}
For a given $f$ in the form of \eqref{equation:f-star-appendix} with all $p_s$ even, and for small constants $\epsilon > 0$ and $\delta > 0$, we choose $D \geq \frac{4\times 50^2 r_{\star}^3 \sum_{s=1}^{r_\star} p_s^5 \norm{\boldsymbol{\beta}_s}_2^{p_s}}{\epsilon^2 \delta}$, and $m \geq \frac{54 r_\star D(1+\log \frac{8}{\delta})}{\epsilon^2} \log \frac{1}{\epsilon}$. Let $\wb_{0, r} \overset{{\rm iid}}{\sim} \normal(\boldsymbol{0}, \Ib_D)$ and $a_{r} \overset{{\rm iid}}{\sim} \textrm{Unif}(\{\pm 1\})$ for $r = 1, \dots, m$, then there exist proper $\{\wb_r^*\}$ such that with probability at least $1 - \delta$, we have
\begin{align*}
\norm{\frac{1}{2\sqrt{m}} \sum_{r=1}^m a_{r} \mathds{1}\{\wb_{0, r}^\top \hb(\xb) \geq 0\} \left((\wb_r^*)^\top \hb(\xb)\right)^2 - f(\xb)}_{L_2} \leq 7 r_\star \epsilon.
\end{align*}
\end{lemma}
\begin{proof}
By definition, $f$ can be written as a sum of polynomials with leading coefficients $\alpha_s$. We partition $m$ neurons into two parts according to the sign of $a_{r}$. We will use the positive part to express those polynomials with positive coefficient $\alpha_s$, and negative part to express those with negative coefficients. We first show for sufficiently large $m$, the number of positive $a_{r}$'s exceeds $\frac{1}{3}m$ with high probability. This follows from the tail bound of i.i.d. binomial random variables. By the Hoeffding's inequality, we have
\begin{align*}
\PP\left(\textrm{Number of positive $a_{r}$} \leq k\right) \leq \exp\left(-2m (1/2 - (k/m)^2)\right).
\end{align*}
Letting $k = \frac{1}{3} m$ and setting $\PP\left(\textrm{Number of positive $a_{r}$} \leq k\right) \leq \delta$, we have $m \geq 2 \log \frac{1}{\delta}$. We denote $\cI_1 = \{1, \dots, m/3\}$ and $\cI_2 = \{m/3+1, \dots, 2m/3\}$. Without loss of generality, we assume $a_{r} = 1$ for $r \in \cI_1$.

The remaining proof is built upon Lemma \ref{lemma:approx_polys}. We choose $D = \frac{50^2 r_{\star}^3 \sum_{s=1}^{r_\star} p_s^5 \norm{\boldsymbol{\beta}_s}_2^{p_s}}{\epsilon^2 \delta}$, so that with probability at least $1-\delta$, there exists $\ab$ with $\norm{\ab^\top \gb(\xb) - \sum_{s=1}^{r_\star} (\bbeta^\top_s \xb)^{p_s/2}}_{L_2} \leq \epsilon$. We further partition $\cI_1$ into $r_\star$ consecutive groups of equal size $m_0$, i.e., $r_\star m_0 = m/3$. Within a group, we aim to approximate $\alpha_s (\boldsymbol{\beta}_s^\top \xb)^{p_s}$ with $\alpha_s > 0$ for some fixed $s \leq r_\star$. Accordingly, we choose 
$\wb_{r}^{s, *} = 2\sqrt{\alpha_s} (3r_{\star})^{1/4} m_0^{-1/4} \hat{\bSigma}^{1/2} [\boldsymbol{0}^\top, \dots, \ab_s^\top, \dots, \boldsymbol{0}^\top]^\top$ for $r = 1, \dots, m_0$. 
We have
\begin{align}
& \norm{\frac{1}{2\sqrt{m}} \sum_{r=1}^{m_0} a_{0, r}\mathds{1}\{\wb_{0, r}^\top \hb(\xb) \geq 0\} \left((\wb_{r}^{s, *})^\top \hb(\xb)\right)^2 - \alpha_s(\boldsymbol{\beta}_s^\top \xb)^{p_s}}_{L_2} \notag \\
=~& \norm{\frac{1}{2\sqrt{3r_{\star}m_0}} \sum_{r=1}^{m_0} 4\sqrt{3r_{\star}}\mathds{1}\{\wb_{0, r}^\top \hb(\xb) \geq 0\} \alpha_s m_0^{-1/2}\left(\ab_s^\top \gb_s\right)^2 - \alpha_s(\boldsymbol{\beta}_s^\top \xb)^{p_s}}_{L_2} \notag \\
=~& \norm{\frac{1}{m_0} \sum_{r=1}^{m_0} 2 \alpha_s \mathds{1}\{\wb_{0, r}^\top \hb(\xb) \geq 0\} \left(\ab_s^\top \gb_s\right)^2 - \alpha_s(\boldsymbol{\beta}_s^\top \xb)^{p_s}}_{L_2}. \label{eq:L2_group}
\end{align}
We know $\ab_s^\top \gb_s$ well approximates $(\boldsymbol{\beta}_s^\top \xb)^{p_s/2}$. If $\sup_{\xb \in \SSS^{d-1}} \frac{1}{m_0} \sum_{r=1}^{m_0} 2 \mathds{1}\{\wb_{0, r}^\top \hb(\xb) \geq 0\}$ concentrates around $1$, then the above $L_2$ norm can be bounded by $O(\epsilon)$. We substantiate this reasoning by the following claim:
\begin{claim}\label{claim:indicator_concentration}
With probability at least $1-2\delta$, we have
\begin{align*}
\sup_{\xb \in \SSS^{d-1}} \left|\frac{1}{m_0} \sum_{r=1}^{m_0} 2 \mathds{1}\{\wb_{0, r}^\top \hb(\xb) \geq 0\} - 1\right| \leq 6\sqrt{\frac{D\log (3m_0) \left(1+\log \frac{2}{\delta}\right)}{m_0}}.
\end{align*}
\end{claim}
The proof of the claim is deferred to Appendix \ref{pf:claim}. Based on the claim,
we are ready to finish proving \eqref{eq:L2_group}. By the triangle inequality, we deduce
\begin{align*}
& \norm{\frac{1}{m_0} \sum_{r=1}^{m_0} 2 \alpha_s \mathds{1}\{\wb_{0, r}^\top \hb(\xb) \geq 0\} \left(\ab_s^\top \gb_s\right)^2 - \alpha_s(\boldsymbol{\beta}_s^\top \xb)^{p_s}}_{L_2} \\
=~& \norm{\frac{1}{m_0} \sum_{r=1}^{m_0} 2 \alpha_s \mathds{1}\{\wb_{0, r}^\top \hb(\xb) \geq 0\} \left(\ab_s^\top \gb_s\right)^2 - \alpha_s \left(\ab_s^\top \gb_s\right)^2 + \alpha_s \left(\ab_s^\top \gb_s\right)^2 - \alpha_s(\boldsymbol{\beta}_s^\top \xb)^{p_s}}_{L_2} \\
\leq~& \norm{\frac{1}{m_0} \sum_{r=1}^{m_0} 2 \alpha_s \mathds{1}\{\wb_{0, r}^\top \hb(\xb) \geq 0\} \left(\ab_s^\top \gb_s\right)^2 - \alpha_s \left(\ab_s^\top \gb_s\right)^2}_{L_2} + \norm{\alpha_s \left(\ab_s^\top \gb_s\right)^2 - \alpha_s(\boldsymbol{\beta}_s^\top \xb)^{p_s}}_{L_2} \\
\leq~& \alpha_s \norm{\left(\ab_s^\top \gb_s\right)^2}_{L_2} \norm{\frac{1}{m_0} \sum_{r=1}^{m_0} 2 \mathds{1}\{\wb_{0, r}^\top \hb(\xb) \geq 0\} - 1}_{L_2} \\
& + \alpha_s \norm{\ab_s^\top \gb_s + (\boldsymbol{\beta}_s^\top \xb)^{p_s/2}}_{L_2} \norm{\ab_s^\top \gb_s - (\boldsymbol{\beta}_s^\top \xb)^{p_s/2}}_{L_2} \\
\leq~& 6 \alpha_s \left(\norm{(\boldsymbol{\beta}_s^\top \xb)^{p_s/2}}_{L_2} + \epsilon\right) \sqrt{\frac{D\log (3m_0) \left(1+\log \frac{2}{\delta}\right)}{m_0}} + \alpha_s \epsilon \left(2\norm{(\boldsymbol{\beta}_s^\top \xb)^{p_s/2}}_{L_2} + \epsilon\right).
\end{align*}
The above upper bound holds with probability no smaller than $1-3\delta$. Taking $$m_0 = \frac{18 D\left(1+\log \frac{2}{\delta}\right)}{\epsilon^2} \log \frac{1}{\epsilon},$$ for a small $\epsilon < \norm{(\boldsymbol{\beta}_s^\top \xb)^{p_s/2}}_{L_2}$, with probability at least $1- 3\delta$, the following
\begin{align*}
\norm{\frac{1}{m_0} \sum_{r=1}^{m_0} a_{r}\mathds{1}\{\wb_{0, r}^\top \hb(\xb) \geq 0\} \left((\wb_{r}^{s, *})^\top \hb(\xb)\right)^2 - \alpha_s(\boldsymbol{\beta}_s^\top \xb)^{p_s}}_{L_2} \leq 7 \alpha_s \epsilon \norm{(\boldsymbol{\beta}_s^\top \xb)^{p_s/2}}_{L_2}
\end{align*}
holds true for the $s$-th group with $\alpha_s > 0$. When $\alpha_s < 0$, we simply set $\wb_r^{s, \star} = \boldsymbol{0}$. As a result, in $\cI_1$, we can express all the polynomial with a positive coefficient. 

To express polynomials with negative coefficients, we use $\cI_2$ analogously. By evenly partitioning $\cI_2$ into $r_\star$ consecutive groups, for a fixed $s \leq r_{\star}$ and $\alpha_s < 0$, we choose $\wb_{r}^{s, *} = 2\sqrt{|\alpha_s|} (3r_{\star})^{1/4} m_0^{-1/4} \hat{\bSigma}^{1/2} [\boldsymbol{0}^\top, \dots, \ab_s^\top, \dots, \boldsymbol{0}^\top]^\top$. Using exactly the same argument in $\cI_1$, with probability at least $1-3\delta$, for $\alpha_s < 0$, we also have
\begin{align*}
\norm{\frac{1}{m_0} \sum_{r=1}^{m_0} a_{r}\mathds{1}\{\wb_{0, r}^\top \hb(\xb) \geq 0\} \left((\wb_{r}^{s, *})^\top \hb(\xb)\right)^2 - \alpha_s(\boldsymbol{\beta}_s^\top \xb)^{p_s}}_{L_2} \leq 7 |\alpha_s| \epsilon \norm{(\boldsymbol{\beta}_s^\top \xb)^{p_s/2}}_{L_2}.
\end{align*}
The last step for proving Lemma \ref{lemma:topexpress} is to combine $\cI_1$ and $\cI_2$ together and choose the remaining weight parameters $\wb_r^*$ identically $\boldsymbol{0}$ for $r \geq 2m/3+1$. 
Substituting into the \quadh~model, with probability at least $1-4\delta$, we deduce
\begin{align*}
& \norm{\frac{1}{2\sqrt{m}} \sum_{r=1}^m a_{r} \mathds{1}\{\wb_{0, r}^\top \hb(\xb) \geq 0\} \left((\wb_r^*)^\top \hb(\xb)\right)^2 - f(\xb)}_{L_2} \\
=~ & \norm{\frac{1}{2\sqrt{m}} \sum_{r \in \cI_1 \bigcup \cI_2} a_{r} \mathds{1}\{\wb_{0, r}^\top \hb(\xb) \geq 0\} \left((\wb_r^*)^\top \hb(\xb)\right)^2 - f(\xb)}_{L_2} \\
\leq~ & \sum_{s=1}^{r_\star} \norm{\frac{1}{2\sqrt{m}} \sum_{r \in \cI_1 \bigcup \cI_2} a_{r}\mathds{1}\{\wb_{0, r}^\top \hb(\xb) \geq 0\} \left((\wb_{r}^{s, *})^\top \hb(\xb)\right)^2 - \alpha_s (\boldsymbol{\beta}_s^\top \xb)^{p_s}}_{L_2} \\
\leq~ & \sum_{s=1}^{r_\star} \norm{\frac{1}{m_0} \sum_{r=1}^{m_0} 2 \alpha_s \mathds{1}\{\wb_{0, r}^\top \hb(\xb) \geq 0\} \left(\ab_s^\top \gb_s(\xb)\right)^2 - \alpha_s(\boldsymbol{\beta}_s^\top \xb)^{p_s}}_{L_2} \\
\leq~ & 7 \epsilon \sum_{s=1}^{r_\star} |\alpha_s| \norm{(\boldsymbol{\beta}_s^\top \xb)^{p_s/2}}_{L_2} \\
\leq~ & 7 \epsilon \sqrt{\sum_{s=1}^{r_\star} \alpha_s^2 \sum_{s=1}^{r_\star} \norm{(\boldsymbol{\beta}_s^\top \xb)^{p_s}}_{L_2}} \\
\leq~ & 7 r_\star \epsilon.
\end{align*}
The width $m$ satisfies $m = 3 r_\star m_0 \geq \frac{54 r_\star D(1+\log \frac{2}{\delta})}{\epsilon^2} \log \frac{1}{\epsilon}$. Replacing $\delta = \delta /4$ completes the proof.
\end{proof}

{\bf Expressivity with odd-degree polynomials}.

\quadh~model can also efficiently express odd-degree polynomials. We rely on the following decomposition trick. Let $k$ be an integer. We rewrite a $(2k+1)$-degree polynomial as
\begin{align*}
(\boldsymbol{\beta}^\top \xb)^{2k+1} = \left(\frac{(\boldsymbol{\beta}^\top \xb)^{k+1} + (\boldsymbol{\beta}^\top \xb)^{k}}{2} \right)^2 - \left(\frac{(\boldsymbol{\beta}^\top \xb)^{k+1} - (\boldsymbol{\beta}^\top \xb)^{k}}{2} \right)^2.
\end{align*}
Since QuadNTK can naturally implement the quadratic function, we only require that the neural representation $\hb(\xb)$ can approximate $(\boldsymbol{\beta}^\top \xb)^{k+1} \pm (\boldsymbol{\beta}^\top \xb)^{k}$. This is true since random indicator functions can approximate $(\boldsymbol{\beta}^\top \xb)^{k+1}$ and $(\boldsymbol{\beta}^\top \xb)^{k}$ due to Lemma \ref{lemma:bottomexpress}. We denote $\ab_1^\top \gb_1(\xb) \approx (\boldsymbol{\beta}^\top \xb)^{k+1}$ in $L_2$, and $\ab_2^\top \gb_2(\xb) \approx (\boldsymbol{\beta}^\top \xb)^{k}$ in $L_2$. Then by stacking $\gb_1$ and $\gb_2$, we have $[\ab_1^\top, \pm\ab_2^\top] [\gb_1^\top, \gb_2^\top]^\top \approx (\boldsymbol{\beta}^\top \xb)^{k+1} \pm (\boldsymbol{\beta}^\top \xb)^{k}$ in $L_2$. Therefore, we only need to augment the dimension $D$ of the neural representation to approximate odd-degree polynomials. We concretize this argument in the following lemma.
\begin{lemma}\label{lemma:odd}
For a given $f$ in the form of \eqref{equation:f-star-appendix}, and small constants $\epsilon > 0$ and $\delta > 0$, we choose $D \geq \frac{8\times 50^2 r_{\star}^3 \sum_{s=1}^{r_\star} p_s^5 \norm{\boldsymbol{\beta}_s}_2^{2\lceil p_s/2 \rceil}}{\epsilon^2 \delta}$, and $m \geq \frac{54 r_\star D(1+\log \frac{8}{\delta})}{\epsilon^2} \log \frac{1}{\epsilon}$. Let $\wb_{0, r} \overset{{\rm iid}}{\sim} \normal(\boldsymbol{0}, \Ib_D)$ and $a_{r} \overset{{\rm iid}}{\sim} \textrm{Unif}(\{\pm 1\})$ for $r = 1, \dots, m$, then there exist proper $\{\wb_r^*\}$ such that with probability at least $1 - \delta$, we have
\begin{align*}
\norm{\frac{1}{2\sqrt{m}} \sum_{r=1}^m a_{r} \mathds{1}\{\wb_{0, r}^\top \hb(\xb) \geq 0\} \left((\wb_r^*)^\top \hb(\xb)\right)^2 - f(\xb)}_{L_2} \leq 7 r_\star \epsilon.
\end{align*}
\end{lemma}
\begin{proof}
We can write
\begin{align*}
f(\xb) & = \sum_{s=1}^{r_\star} \alpha_s (\boldsymbol{\beta}_s^\top \xb)^{p_s} \ind\{p_s ~\textrm{is even}\} + \alpha_s (\boldsymbol{\beta}_s^\top \xb)^{p_s} \ind\{p_s ~\textrm{is odd}\} \\
& = \sum_{s=1}^{r_\star} \alpha_s \left((\boldsymbol{\beta}_s^\top \xb)^{p_s/2}\right)^2 \ind\{p_s ~\textrm{is even}\} \\
& \quad + \alpha_s\left[\left(\frac{(\boldsymbol{\beta}_s^\top \xb)^{\frac{p_s+1}{2}} + (\boldsymbol{\beta}_s^\top \xb)^{\frac{p_s-1}{2}}}{2} \right)^2 - \left(\frac{(\boldsymbol{\beta}_s^\top \xb)^{\frac{p_s+1}{2}} - (\boldsymbol{\beta}_s^\top \xb)^{\frac{p_s-1}{2}}}{2}\right)^2\right] \ind\{p_s ~\textrm{is odd}\}.
\end{align*}
Applying Lemma \ref{lemma:bottomexpress} once, there exists $\ab_s$ such that $\norm{\ab_s^\top - (\boldsymbol{\beta}_s^\top \xb)^{p_s/2}}_{L_2} \leq \epsilon/r_\star$, when $p_s$ is even and the corresponding $D_s \geq \frac{4 \times 50^2 r_{\star}^3 \sum_{s=1}^{r_\star}  p_s^5 \norm{\boldsymbol{\beta}_s}_2^{2p_s}}{\epsilon^2 \delta}$. 

For an odd $p_s$, we apply the technique in Lemma \ref{lemma:Dbound}. There exist $\ab_{s, +}$ and $\ab_{s, -}$ with corresponding random indicator features $\gb_{s, +}(\xb)$ and $\gb_{s, -}(\xb)$ such that 
\begin{align*}
& \norm{[\ab_{s, +}^\top, \pm \ab_{s, -}^\top] [\gb_{s, +}^\top(\xb), \gb_{s, -}^\top(\xb)]^\top - \left((\boldsymbol{\beta}_s^\top \xb)^{\frac{p_s+1}{2}} \pm (\boldsymbol{\beta}_s^\top \xb)^{\frac{p_s-1}{2}} \right)}_{L_2} \\
\leq~ & \norm{\ab_{s, +}^\top \gb_{s, +}(\xb) - (\boldsymbol{\beta}_s^\top \xb)^{p_s/2}}_{L_2} + \norm{\ab_{s, -}^\top \gb_{s, -}(\xb) - (\boldsymbol{\beta}_s^\top \xb)^{\frac{p_s-1}{2}}}_{L_2} \\
\leq~ & \epsilon/r_{\star}.
\end{align*}
The corresponding neural representation dimension is $D_s \geq 50^2 r_{\star}^3 \frac{(p_s+1)^5 \norm{\boldsymbol{\beta}_s}_2^{p_s+1} + (p_s-1)^5 \norm{\boldsymbol{\beta}_s}_2^{p_s-1}}{\epsilon^2 \delta}$.
Combining the even and odd degrees together, we can choose
\begin{align*}
D \geq \frac{4\times 50^2 r_{\star}^3}{\epsilon^2 \delta} \left(\sum_{s=1}^{r_\star} p_s^5 \norm{\boldsymbol{\beta}_s}_2^{p_s} \ind\{p_s\textrm{~is even}\} + 2(p_s+1)^5 \norm{\boldsymbol{\beta}_s}_2^{p_s+1} \ind\{p_s\textrm{~is odd}\}\right).
\end{align*}
Lemma \ref{lemma:odd} now follows from Lemma \ref{lemma:topexpress} by merging $\gb_{s, +}, \gb_{s, -}$ as a single feature $\gb_s$, and $\ab_{s, +}, \ab_{s, -}$ as a single weight vector $\ab_s$ so that $\wb_r^{s, *}$ can be chosen accordingly. Unifying the notation for even and odd degree polynomials, we have
\begin{align*}
D \geq \frac{8\times50^2 r_{\star}^3}{\epsilon^2 \delta} \left(\sum_{s=1}^{r_\star} (p_s+1)^5 \norm{\boldsymbol{\beta}_s}_2^{2 \lceil p_s/2\rceil}\right).
\end{align*}
\end{proof}

\subsection{Generalization of \quadh}\label{quadneuralgen}
Lemma \ref{lemma:topexpress} and \ref{lemma:odd} construct a proper weight matrix $\Wb^* = [\wb_1, \dots, \wb_m]^\top$ such that \quadneural~can well approximates $f_\star$ of form \eqref{equation:f-star-appendix} in the $L_2$ sense. Now we show that for sufficiently large $m$, the empirical risk $\hat{\cR}_{\ell}(f_{\Wb^*})$ is comparable to that of $f_\star$. By the Lipschitz property of the loss function, we derive
\begin{align*}
\frac{1}{n} \sum_{i=1}^n \ell(f_{\Wb^*}(\xb_i), y_i) & \leq \frac{1}{n} \sum_{i=1}^n \ell(f_{\Wb^*}(\xb_i), y_i) - \ell(f_\star(\xb_i), y_i) + \ell(f_0(\xb_i), y_i) \\
& \leq \frac{1}{n} \sum_{i=1}^n \left|f_{\Wb^*}(\xb_i) - f_\star(\xb_i)\right| + \hat{\cR}(f_\star).
\end{align*}
For a given $\epsilon_0 > 0$, using Chebyshev's inequality, we have
\begin{align*}
\PP\left(|\hat{\cR}(f_\star) - \cR(f_\star) | \geq \epsilon_0/2\right) & \leq \frac{4 \EE[(\hat{\cR}(f_\star) - \cR(f_\star))^2]}{\epsilon_0^2} \\
& = \frac{4\EE_{(\xb, y) \sim \cD}[\ell(f_\star(\xb), y) - \EE_{(\xb, y) \sim \cD} [\ell(f_\star(\xb), y)]]^2}{n_2 \epsilon_0^2} \\
& \leq \frac{4\EE_{(\xb, y) \sim \cD}[\ell(0, y) + |f_\star(\xb)|- \mathsf{OPT}]^2}{n \epsilon_0^2} \\
& \leq \frac{8\EE_{(\xb, y) \sim \cD}[|f_\star(\xb)|]^2 + 8(1+ \mathsf{OPT})^2}{n \epsilon_0^2} \\
& \leq \frac{8(1+ \mathsf{OPT})^2 + 8 \EE_{\xb} \left[\sum_{s=1}^{r_\star} \alpha_s^2 \sum_{s=1}^{r_\star} (\boldsymbol{\beta}_s^\top \xb)^{2p_s}\right]}{n \epsilon_0^2} \\
& \leq \frac{8(1+ \mathsf{OPT})^2 + 8 r_\star^2}{n \epsilon_0^2}.
\end{align*}
Choosing $n \geq \frac{8(1+ \mathsf{OPT})^2 + 8 r_\star^2}{\delta \epsilon_0^2}$, $\hat{\cR}(f_\star) - \cR(f_\star) \leq \epsilon_0/2$ holds with probability at least $1-\delta$. We further invoke Lemma \ref{lemma:topexpress} and Chebyshev's inequality again on $\frac{1}{n} \sum_{i=1}^n \left|f_{\Wb^*}(\xb_i) - f_\star(\xb_i)\right|$:
\begin{align*}
\PP_{\xb}\left(\frac{1}{n} \sum_{i=1}^n \left|f_{\Wb^*}(\xb_i) - f_\star(\xb_i)\right| \geq \epsilon_0/2\right) & \leq \frac{4\EE_{\xb} \left[\frac{1}{n^2} \left(\sum_{i=1}^n |f_{\Wb^*}(\xb_i) - f_\star(\xb_i)|\right)^2\right]}{\epsilon_0^2} \\
& \leq \frac{\frac{4}{n} \sum_{i=1}^n \EE_{\xb} \left[|f_{\Wb^*}(\xb_i) - f_\star(\xb_i)|^2 \right]}{\epsilon_0^2} \\
& \leq \frac{196 \epsilon^2 r_\star^2}{\epsilon_0^2},
\end{align*}
where the last inequality holds with probability at least $1-\delta$. We set $\frac{196 r_\star^{2} \epsilon^2}{\epsilon_0^2} \leq \delta$, which implies $\epsilon^2 \leq \frac{\delta \epsilon_0^2}{196 r_\star^2}$. Accordingly, the number of neurons in the top layer needs to be at least $$m \geq \frac{10584 r_\star^3 D\left(1+\log \frac{8}{\delta}\right) \sum_{s=1}^{r_\star} \norm{(\boldsymbol{\beta}_s^\top \xb)^{p_s}}_{L_2}}{\delta \epsilon_0^2} \log \frac{7r_\star}{\sqrt{\delta}\epsilon_0},$$ and the dimension of the neural representation is 
\begin{align}\label{eq:Dsize}
D = \frac{50^2 \times 392 r_{\star}^5}{\delta \epsilon_0^2} \left(\sum_{s=1}^{r_\star} (p_s+1)^5 \norm{\boldsymbol{\beta}_s}_2^{2 \lceil p_s/2\rceil}\right).
\end{align}
This gives us that with probability at least $1-3\delta$ over the randomness of data and initialization\footnote{To achieve probability $1-\delta$, we replace $\delta$ with $\delta/3$, which only introduce a multiplicative constant in the size of $m$ and $D$.}, the empirical risk satisfies
\begin{align*}
\frac{1}{n} \sum_{i=1}^n \ell(f_{\Wb^*}(\xb_i), y_i) \leq \mathsf{OPT} + \epsilon_0.
\end{align*}
Applying Theorem \ref{thm:opt-gen} part (2), for any second-order stationary point $\hat{\Wb}$ and proper regularization parameter $\lambda$, we have
\begin{align*}
\hat{\cR}_{\lambda}(f^Q_{\hat{\Wb}}) \leq (1+\tau) (\mathsf{OPT}+\epsilon_0) + \epsilon_0 \leq (1+\tau_0) \mathsf{OPT} + \epsilon_0.
\end{align*}

{\bf Bounding $B_{w, \star}$}.

Towards establishing the generalization bound of $f^Q_{\hat{\Wb}}$, we first find $B_{w, \star}$:
\begin{align*}
\sum_{r=1}^m \norm{\wb_{r}^*}_2^4 = \sum_{s=1}^{r_\star} \sum_{r \in \cI_1 \bigcup \cI_2} \norm{\wb_{r}^{s, *}}_2^4 & = 48 r_\star \sum_{s=1}^{r_\star} \sum_{r=1}^{m_0} \alpha_s^2 m_0^{-1}\norm{\hat{\bSigma}^{1/2} [\boldsymbol{0}^\top, \dots, \ab_s^\top, \dots, \boldsymbol{0}^\top]^\top}_2^4 \\
& = 48 r_\star \sum_{s=1}^{r_\star} \alpha_s^2 \norm{\hat{\bSigma}^{1/2} [\boldsymbol{0}^\top, \dots, \ab_s^\top, \dots, \boldsymbol{0}^\top]^\top}_2^4.
\end{align*}
To bound $\norm{\hat{\bSigma}^{1/2} [\boldsymbol{0}^\top, \dots, \ab_s^\top, \dots, \boldsymbol{0}^\top]^\top}_2$, we first replace $\hat{\bSigma}$ with $\bSigma$. We denote $\boldsymbol{\theta}_s = \bSigma^{1/2} [\boldsymbol{0}^\top, \dots, \ab_s^\top, \dots, \boldsymbol{0}^\top]^\top$, and observe $\boldsymbol{\theta}_s$ is the optimal solution to the following least square problem
\begin{align*}
\boldsymbol{\theta}_s = \argmin_{\ub_1} \norm{F(\xb) - \ub_1^\top \boldsymbol{\Sigma}^{-1/2}\gb(\xb)}_{L_2}^2 \quad \textrm{with} \quad F(\xb) = \ab_s^\top \gb_s^\top(\xb).
\end{align*}
The optimal solution is $\boldsymbol{\theta}_s = \ub_1^* = \boldsymbol{\Sigma}^{-1/2} \EE_{\xb} [F(\xb) \gb(\xb)]$. Similarly, the optimal solution to the following least square problem
\begin{align*}
\min_{\ub_2} \norm{F(\xb) - \ub_2^\top \gb(\xb)}_{L_2}^2 \quad \textrm{with} \quad F(\xb) = \ab_s^\top \gb_s(\xb)
\end{align*}
is $\ub^*_{2} = \boldsymbol{\Sigma}^{-1} \EE_{\xb} [F(\xb) g(\xb)]$. The residual of $\ub^*_2$ is
\begin{align*}
\norm{F(\xb) - (\ub_2^*)^\top \gb(\xb)}_{L_2}^2 = \norm{F(\xb)}_{L_2}^2 - \EE_{\xb} [F(\xb) \gb(\xb)^\top] \boldsymbol{\Sigma}^{-1} \EE_{\xb} [F(\xb) \gb(\xb)] \geq 0.
\end{align*}
This implies
\begin{align*}
\norm{\boldsymbol{\theta}_s}_2 = \sqrt{\EE_{\xb} [F(\xb) \gb(\xb)^\top] \boldsymbol{\Sigma}^{-1} \EE_{\xb} [F(\xb) \gb(\xb)]} \leq \norm{F(\xb)}_{L_2} \leq \norm{(\boldsymbol{\beta}_s^\top \xb)^{p_s/2}}_{L_2} + \epsilon,
\end{align*}
where the last inequality follows from Lemma \ref{lemma:bottomexpress}. This gives rise to
\begin{align*}
\sum_{s=1}^{r_\star} \alpha_s^2 \norm{\boldsymbol{\theta}_s}_2^4 \leq \sum_{s=1}^{r_\star} \alpha_s^2 \norm{(\boldsymbol{\beta}^\top \xb)^{p/2}}_{L_2}^4 \leq r_\star.
\end{align*}
To switch back to $\hat{\bSigma}$, we invoke Lemma \ref{lemma:covariance_relative} on the concentration of $\hat{\bSigma}$ to $\bSigma$. Specifically, with probability at least $1-\delta$, choosing $n_0 \geq 4c \delta^{-2} \lambda_{\lceil p/2 \rceil}^{-1} D \log D$ for some constant $c$, we have
\begin{align*}
\frac{1}{2}\boldsymbol{\Sigma} \preceq \hat{\boldsymbol{\Sigma}} \preceq \frac{3}{2} \boldsymbol{\Sigma}.
\end{align*}
Consequently, by denoting $\hat{\boldsymbol{\theta}}_s = \hat{\bSigma}^{1/2} [\boldsymbol{0}^\top, \dots, \ab_s^\top, \dots, \boldsymbol{0}^\top]^\top$, we have
\begin{align*}
\norm{\hat{\boldsymbol{\theta}}_s}_2^2 = \norm{\hat{\boldsymbol{\Sigma}}^{1/2} \boldsymbol{\Sigma}^{-1/2} \boldsymbol{\theta}_s}_2^2 = \boldsymbol{\theta}_s^\top \boldsymbol{\Sigma}^{-1/2} \hat{\boldsymbol{\Sigma}} \boldsymbol{\Sigma}^{-1/2} \boldsymbol{\theta}_s^\top \leq \norm{\boldsymbol{\Sigma}^{-1/2} \hat{\boldsymbol{\Sigma}} \boldsymbol{\Sigma}^{-1/2}}_{\textrm{op}} \norm{\boldsymbol{\theta}_s}_2^2 \leq \frac{3}{2} \norm{\boldsymbol{\theta}_s}_2^2.
\end{align*}
Plugging into $\sum_{r=1}^m \norm{\wb_{r}^*}_2^4$, we have
\begin{align*}
\sum_{r=1}^m \norm{\wb_{r}^*}_2^4 \le 108r_\star \sum_{s=1}^{r_\star} \alpha_s^2 \norm{\boldsymbol{\theta}_s}_2^4 \leq 108r_\star^2.
\end{align*}
Therefore, we can set $B_{w, \star}^4 = 108r_\star^2$. Note that $B_{w, \star}$ is independent of the width $m$.

{\bf Bounding $M_{h, \textrm{op}}$ and $B_h$}.

The remaining ingredients are $M_{h, \textrm{op}}$ and $\norm{\hb(\xb)}_2$. Conditioned on the event $\frac{1}{2}\boldsymbol{\Sigma} \leq \hat{\boldsymbol{\Sigma}} \leq \frac{3}{2} \boldsymbol{\Sigma}$, we know $\hat{\boldsymbol{\Sigma}}^{-1} \leq 2\boldsymbol{\Sigma}^{-1}$. Therefore, we have
\begin{align*}
\norm{\hb(\xb)}_2^2 = \gb(\xb)^\top \hat{\bSigma}^{-1} \gb(\xb) \leq 2 \gb(\xb)^\top \bSigma^{-1} \gb(\xb) \leq 2\lambda_{\lceil p/2 \rceil}^{-1} D.
\end{align*}
Note that he norm of $\hb(\xb)$ is in the order of $\sqrt{D}$ according to Assumption \ref{assumption_covariance}.

Lastly, we bound $M_{h, \textrm{op}}$ as
\begin{align*}
B_h^2 M_{h, \textrm{op}}^2 & = \EE_{\xb} \left[\norm{\frac{1}{n_2} \sum_{i=n_1+1}^n \hb(\xb_i) \hb(\xb_i)^\top}_{\rm op} \right] \\
& = \EE_{\xb} \left[\norm{\hat{\boldsymbol{\Sigma}}^{-1/2} \left(\frac{1}{n_2} \sum_{i=n_1+1}^n \gb(\xb_i) \gb(\xb_i)^\top\right) \hat{\boldsymbol{\Sigma}}^{-1/2}}_{\rm op} \right] \\
& \leq \EE_{\xb} \left[\norm{\hat{\boldsymbol{\Sigma}}^{-1/2} \bSigma^{1/2} \bSigma^{-1/2} \left(\frac{1}{n_2} \sum_{i=n_1+1}^n \gb(\xb_i) \gb(\xb_i)^\top\right) \bSigma^{-1/2} \bSigma^{1/2} \hat{\boldsymbol{\Sigma}}^{-1/2}}_{\rm op}\right] \\
& \leq \norm{\hat{\boldsymbol{\Sigma}}^{-1/2} \bSigma^{1/2}}_{\rm op} \EE_{\xb}\left[\norm{\bSigma^{-1/2} \left(\frac{1}{n_2} \sum_{i=n_1+1}^n \gb(\xb_i) \gb(\xb_i)^\top\right) \bSigma^{-1/2}}_{\rm op}\right] \norm{\bSigma^{1/2} \hat{\boldsymbol{\Sigma}}^{-1/2}}_{\rm op} \\
& \leq \frac{3}{2} \norm{\hat{\boldsymbol{\Sigma}}^{-1/2} \bSigma^{1/2}}_{\rm op}^2.
\end{align*}
The last inequality holds, due to Lemma \ref{lemma:covariance_relative} and $\hat{\bSigma}$ is obtained using independent samples. Conditioned on the same event $\frac{1}{2}\boldsymbol{\Sigma} \leq \hat{\boldsymbol{\Sigma}} \leq \frac{3}{2} \boldsymbol{\Sigma}$, we have
\begin{align*}
\norm{\hat{\boldsymbol{\Sigma}}^{-1/2} \bSigma^{1/2}}_{\rm op}^2 = \norm{\bSigma^{1/2}\hat{\bSigma}^{-1}\bSigma^{1/2}}_{\rm op} \leq 2.
\end{align*}
Therefore, $M_{h, \textrm{op}}^2 \leq 3 B_h^{-2}$. Putting all the ingredients together and applying Theorem \ref{thm:opt-gen}, by choosing $$m \geq \max\set{\frac{10584 r_\star^3 D\left(1+\log \frac{8}{\delta}\right) \sum_{s=1}^{r_\star} \norm{(\boldsymbol{\beta}_s^\top \xb)^{p_s}}_{L_2}}{\delta \epsilon_0^2} \log \frac{7r_\star}{\sqrt{\delta}\epsilon_0}, ~\frac{108C^2 D^2 r_\star^2}{\epsilon_0 \sqrt{2\lambda_0}}},$$ we establish for any SOSP $\hat\Wb$, the generalization error bounded by:
\begin{align*}
\EE_{(\xb_i, y_i)}\left[\left|\cR(f^Q_{\hat{\Wb}})- \hat{\cR}(f^Q_{\hat{\Wb}})\right|\right] & \leq \tilde{O}\left(\frac{\norm{\hb(\xb)}_2^2 B_{w, \star}^2 M_{h, \textrm{op}}}{\sqrt{n}}\right) = \tilde{O} \left(\sqrt{\frac{2\lambda_{\lceil p/2 \rceil}^{-1} D r_{\star}^2}{n}}\right).
\end{align*}
Using Markov's inequality, we have
\begin{align*}
\PP\left(\left|\cR(f^Q_{\hat{\Wb}})- \hat{\cR}(f^Q_{\hat{\Wb}})\right| \geq \epsilon_0 \right) & \leq \frac{\EE_{(\xb_i, y_i)} \left[\left|\cR(f^Q_{\hat{\Wb}})- \hat{\cR}(f^Q_{\hat{\Wb}})\right|\right]}{\epsilon_0} \\
& \leq \tilde{O}\left(\epsilon_0^{-1}\sqrt{\frac{2\lambda_{\lceil p/2 \rceil}^{-1} D r_{\star}^2}{n}}\right).
\end{align*}
We set the above probability upper bounded by $\delta$, which requires
\begin{align*}
n = \tilde{O}\left(\frac{\lambda_{\lceil p/2 \rceil}^{-1} r_{\star}^7}{\epsilon_0^4 \delta^3} \left(\sum_{s=1}^{r_\star} (p_s+1)^5\norm{\boldsymbol{\beta}_s}_2^{2\lceil p_s/2\rceil}\right) \right).
\end{align*}
We can now bound $\cR(f^Q_{\hat{W}})$ as
\begin{align*}
\cR(f^Q_{\hat{\Wb}}) & = \cR(f^Q_{\hat{\Wb}}) - \hat\cR(f^Q_{\hat{\Wb}}) + \hat\cR(f^Q_{\hat{\Wb}}) \leq (1+\tau_0) \mathsf{OPT} + 2\epsilon_0,
\end{align*}
which holds with probability at least $1-\delta$.

Taking $\norm{\boldsymbol{\beta}_s}_2 = \sqrt{d}$, the sample size $n$ grows in the order of $\tilde{O}\left(\frac{d^{\lceil p/2\rceil}}{\epsilon_0^4} \frac{\lambda_{\lceil p/2 \rceil}^{-1} r_{\star}^{8} p^5}{\delta^3}\right)$. On the other hand, estimating covariance matrix $\bSigma$ requires $n_0 = \tilde{O}\left(4 \delta^{-2} \lambda_{\lceil p/2 \rceil}^{-1} D \log D \right)$ samples, which is in the order of $\tilde{O}\left(\frac{d^{\lceil p/2 \rceil}}{\epsilon_0^2} \frac{ \lambda_{\lceil p/2 \rceil}^{-1} r_{\star}^{6} p^5}{\delta^2}\right)$. Adding $n_1, n_2$ together, the sample complexity $n$ grows in the order of $\tilde{O}\left(\frac{d^{\lceil p/2\rceil}}{\epsilon_0^4} \textrm{poly}(r_\star, p, \delta^{-1})\right)$.

\subsection{Proof of Claim \ref{claim:indicator_concentration}}\label{pf:claim}
\begin{proof}
To show $\sup_{\xb \in \SSS^{d-1}} \frac{1}{m_0} \sum_{r=1}^{m_0} 2 \mathds{1}\{\wb_{0, r}^\top \hb(\xb) \geq 0\}$ is well concentrated, we observe that by symmetry, the following holds
\begin{align*}
\sup_{\xb \in \SSS^{d-1}} \frac{1}{m_0} \sum_{r=1}^{m_0} 2 \mathds{1}\{\wb_{0, r}^\top \hb(\xb) \geq 0\} & = \sup_{\xb \in \SSS^{d-1}} \frac{1}{m_0} \sum_{r=1}^{m_0} 2 \mathds{1}\{\wb_{0, r}^\top \hb(\xb) / \norm{\hb(\xb)}_2 \geq 0\} \\
& \leq \sup_{\yb \in \SSS^{D-1}} \frac{1}{m_0} \sum_{r=1}^{m_0} 2 \mathds{1}\{\wb_{0, r}^\top \yb \geq 0\}.
\end{align*}
For a given $\yb$, each $2 \ind\{\wb_{0, r}^\top \yb \geq 0\}$ is bounded in $[0, 2]$, hence it is sub-Gaussian with variance proxy $1$. Using the Hoeffding's inequality, for every $\yb$, we have
\begin{align*}
\PP\left(\left|\frac{1}{m_0} \sum_{r=1}^{m_0} 2\ind\{\wb_{0, r}^\top \yb\geq 0\} - 1\right| \geq t \right) \leq 2\exp\left(- m_0 t^2 / 2\right).
\end{align*}
To bound the supremum, we discretize the unit sphere. Let $\{\bar{\yb}_i\}_{i=1}^{\normal(\gamma, \SSS^{D-1}, \norm{\cdot}_2)}$ be a $\gamma$-covering of $\SSS^{D-1}$ with $\gamma < 1$, where $\cN(\gamma, \SSS^{D-1}, \norm{\cdot}_2)$ denotes the covering number. By the volume ratio argument, we bound $\cN(\gamma, \SSS^{D-1}, \norm{\cdot}_2) \leq \left(\frac{3}{\gamma}\right)^{D}$. Applying the union bound, we derive
\begin{align*}
\PP\left( \max_{\yb \in \{\bar{\yb}_i\}_{i=1}^{\cN(\gamma, \SSS^{D-1}, \norm{\cdot}_2)}} \left|\frac{1}{m_0} \sum_{r=1}^{m_0} 2\ind\{\wb_{0, r}^\top \yb \geq 0\} - 1\right| \geq t \right) &\leq 2\cN(\gamma, \SSS^{D-1}, \norm{\cdot}_2) \exp\left(- m_0 t^2/2\right) \\
& \leq 2\exp\left(- m_0 t^2 + D \log \frac{3}{\gamma} \right).
\end{align*}
Taking $t = \sqrt{\frac{D \log \frac{3}{\gamma} \left(1+\frac{1}{D}\log \frac{2}{\delta}\right)}{m_0}}$, with probability at least $1-\delta$, we have
\begin{align}\label{eq:coveringmax}
\max_{\yb \in \{\bar{\yb}_i\}_{i=1}^{\normal(\gamma, \SSS^{D-1}, \norm{\cdot}_2)}} \left|\frac{1}{m_0} \sum_{r=1}^{m_0} 2 \ind\{\wb_{0, r}^\top \yb \geq 0\} - 1\right| \leq \sqrt{\frac{D \log \frac{3}{\gamma} \left(1+\frac{1}{D}\log \frac{2}{\delta}\right)}{m_0}}.
\end{align}
By the definition of $\gamma$-covering, for any given $\yb \in \SSS^{D-1}$, there exists $\bar{\yb}$ such that $\norm{\bar{\yb} - \yb}_2 \leq \gamma$. We evaluate how many pairs $\ind\{\wb_{0, r}^\top \yb \geq 0\}, \ind\{\wb_{0, r}^\top \bar{\yb} \geq 0\})$ taking different values, which is equivalent to $(\wb_{0, r}^\top \yb, \wb_{0, r}^\top \bar{\yb})$ having opposite signs. Using the Hoeffding's inequality again, with probability at least $1-\delta$, we have
\begin{align*}
& \Bigg|\frac{1}{m_0}\sum_{r=1}^{m_0} \ind\{\wb_{0, r}^\top \yb, \wb_{0, r}^\top \bar{\yb} \textrm{~having opposite signs}\} \\
&\qquad  - \EE\left[\frac{1}{m_0}\sum_{r=1}^{m_0} \ind\{\wb_{0, r}^\top \yb, \wb_{0, r}^\top \bar{\yb} \textrm{~having opposite signs}\}\right] \Bigg| \leq \sqrt{\frac{\log (2/\delta)}{2m_0}}.
\end{align*}
To bound the expectation, we observe that $(\wb_{0, r}^\top \yb, \wb_{0, r}^\top \bar{\yb})$ is jointly Gaussian with zero mean and the covariance matrix $$\begin{pmatrix}1 & \yb^\top \bar{\yb} \\ \yb^\top \bar{\yb} & 1 \end{pmatrix}.$$ Therefore, we find the following probability
\begin{align*}
\PP&\left(\wb_{0, r}^\top \yb, \wb_{0, r}^\top \bar{\yb} \textrm{~opposite signs}\right)  = 2 \PP\left(\wb_{0, r}^\top \yb \geq 0, \wb_{0, r}^\top \bar{\yb} \leq 0 \right) \\
&= 2 \int_{0}^\infty \int_{-\infty}^0 \frac{1}{2\pi} \left(1 - (\yb^\top \bar{\yb})^2\right)^{-1/2} \exp\left(-\frac{u^2 - (\yb^\top \bar{\yb}) uv + v^2}{2(1 - (\yb^\top \bar{\yb})^2)}\right) dudv \\
& \overset{(i)}{\leq} 2 \int_{0}^\infty \int_{-\infty}^0 \frac{1}{2\pi} \left(1-(\yb^\top \bar{\yb})^2\right)^{-1/2} \exp\left(-\frac{u^2 + v^2}{2(1 - (\yb^\top \bar{\yb})^2)}\right) dudv \\
& = 2 \int_{0}^\infty \frac{1}{\sqrt{2\pi}} \exp\left(-\frac{v^2}{2(1 - (\yb^\top \bar{\yb})^2)}\right) dv \\
& = \sqrt{1 - (\yb^\top \bar{\yb})^2},
\end{align*}
where inequality $(i)$ holds since $uv < 0$. We further bound $1 - (\yb^\top \bar{\yb})^2 = 1 - (1 + \yb^\top (\bar{\yb}-\yb))^2 \leq 1 - (1-\gamma)^2 \leq 2\gamma$. Consequently, we deduce $\PP\left(\wb_{0, r}^\top \yb, \wb_{0, r}^\top \bar{\yb} \textrm{~having opposite signs}\right) \leq \sqrt{2\gamma}$. Taking $\gamma = m_0^{-1}\log 1/\delta$, we have
\begin{align*}
\EE\left[\frac{1}{m_0}\sum_{r=1}^{m_0} \ind\{\wb_{0, r}^\top \yb, \wb_{0, r}^\top \bar{\yb} \textrm{~having opposite signs}\}\right] \leq \sqrt{\frac{2\log (2/\delta)}{m_0}}.
\end{align*}
This implies with probability at least $1-\delta$,
\begin{align}\label{eq:deviation}
\sqrt{\frac{\log (2/\delta)}{2m_0}}\leq \frac{1}{m_0}\sum_{r=1}^{m_0} \ind\{\wb_{0, r}^\top \yb, \wb_{0, r}^\top \bar{\yb} \textrm{~having opposite signs}\} \leq \sqrt{\frac{9\log (2/\delta)}{2m_0}}.
\end{align}
Combining \eqref{eq:coveringmax} and \eqref{eq:deviation} together, with probability at least $1-2\delta$, we deduce
\begin{align*}
 \sup_{\yb \in \SSS^{d-1}} &\left|\frac{1}{m_0}\sum_{r=1}^{m_0} 2 \ind\{\wb_{0, r}^\top \yb \geq 0\} - 1\right| \\&
\leq~ \sup_{\norm{\yb - \bar{\yb}}_2 \leq m_0^{-1}\log 1/\delta}\left| \frac{1}{m_0}\sum_{r=1}^{m_0} 2 \ind\{\wb_{0, r}^\top \yb, \wb_{0, r}^\top \bar{\yb} \textrm{~having opposite signs}\} \right| \\
& + \max_{\yb \in \{\bar{\yb}_i\}_{i=1}^{\normal(\gamma, \SSS^{D-1}, \norm{\cdot}_2)}} \left|\frac{1}{m_0}\sum_{r=1}^{m_0} 2 \ind\{\wb_{0, r}^\top \yb \geq 0\} - 1 \right| \\
\leq~& \sqrt{\frac{9\log \frac{2}{\delta}}{2m_0}} + \sqrt{\frac{D \log(3m_0) \left(1+\frac{1}{D}\log \frac{2}{\delta}\right)}{m_0}} \\
\leq~& 6\sqrt{\frac{D\log (3m_0) \left(1+\log \frac{2}{\delta}\right)}{m_0}}.
\end{align*}
As a result, we know
\begin{align*}
\sup_{\xb \in \SSS^{d-1}} \left|\frac{1}{m_0} \sum_{r=1}^{m_0} 2 \mathds{1}\{\wb_{0, r}^\top \hb(\xb) \geq 0\} - 1\right| \leq 6\sqrt{\frac{D\log (3m_0) \left(1+\log \frac{2}{\delta}\right)}{m_0}}
\end{align*}
holds with probability at least $1-2\delta$. 
\end{proof}

\subsection{Learning in \quadneural~with data dependent regularizer}\label{datadependentreg}
We consider using data dependent regularizer for learning with unwhitened features $\gb(\xb)$, which also yields improved sample complexity. The full learning algorithm is described Algorithm \ref{alg:unwhiten}.
\begin{algorithm}[h]
\caption{Learning with Unwhitened Neural Random Features}
\label{alg:unwhiten}
\begin{algorithmic}
  \STATE {\bf Input}: Labeled data $S_n$, unlabeled data $\wt{S}_{n_0}$, initializations $\Vb\in\R^{D\times d}$, $\bbb\in\R^D$, $\Wb_0\in\R^{m\times D}$, parameters $(\lambda, \epsilon)$.
  \STATE {\bf Step 1:} 1) Construct model $f^Q_\Wb$ as
\leqnomode
  \begin{align*}
\tag*{({\tt Quad-}$\gb$)}& \qquad\qquad
  f^Q_\Wb(\xb) = \frac{1}{2\sqrt{m}}\sum_{r=1}^m a_r\phi''(\wb_{0,r}^\top \gb(\xb)) (\wb_r^\top\gb(\xb))^2,
  \end{align*}
\reqnomode
  where $\gb(\xb)=[\ind\{\vb_1^\top \xb + b_1 \geq 0, \dots, \vb^\top_D \xb + b_D \geq 0\}]^\top $ is the neural random features.
  \STATE 2) Use $\wt{S}_{n_0}$ to estimate the covariance matrix of $\gb(\xb)$, i.e., $\hat\bSigma = \frac{1}{n_0} \sum_{i=1}^{n_0} \gb(\xb_i) \gb(\xb_i)^\top$.
  \STATE {\bf Step 2:} Find a second-order stationary point $\hat{\Wb}$ of the data dependent regularized empirical risk (on the data $S_n$):
  \begin{align*}
    \hat{\cR}^{\rm dreg}_{\lambda}(f^Q_{\Wb}) = \frac{1}{n} \sum_{i=1}^n \ell(f^Q_{\Wb}(\xb_i), y_i) + \lambda \tfnorm{\Wb \hat\bSigma^{1/2}}^4.
  \end{align*}
\end{algorithmic}
\end{algorithm}

Note that \quadg~shares the same QuadNTK model as \quadh, and only replaces the neural representation $\hb$ with $\gb$. The superscript on $\hat\cR_\lambda^{\rm dreg}$ stands for data dependent regularization. We show \quadg~enjoys a similarly nice optimization landscape and good generalization properties as \quadh.

\renewcommand{\thetheorem}{\arabic{theorem}$^\prime$}
\setcounter{theorem}{0}
\begin{theorem}[Optimization landscape and generalization of \quadg]\label{thm:opt-gen-prime}
  Suppose Assumption~\ref{assumption_covariance} holds.
  \begin{enumerate}[wide,label=(\arabic*)]
  \item (Optimization)
    Given any $\epsilon > 0$ and $\delta > 0$, $\tau =\Theta(1)$, and some radius $B_{w,\star}>0$, suppose the width $m\ge \wt{O}(D^2 B_{w,\star}^4\epsilon^{-1})$, sample size $n_0 = \tilde{O}(\delta^{-2} D)$, and we choose a proper regularization coefficient $\lambda>0$. Then with probability $1-\delta$ over $\tilde{S}_{n_0}$, any second-order stationary point $\hat{\Wb}$ of the regularized risk $\hat{\cR}^{\rm dreg}_\lambda(f^Q_\Wb)$ satisfies $\|\hat{\Wb}\hat\bSigma^{1/2}\|_{2, 4} \leq O(B_{w, \star})$, and achieves
    \begin{align*}
\hat{\cR}_{\lambda}^{\rm dreg}(f^Q_{\hat{\Wb}}) \leq (1+\tau) \min_{\tfnorm{\Wb\hat\bSigma^{1/2}} \le B_{w,\star}} \hat{\cR}(f^Q_\Wb) + \epsilon.
    \end{align*}
\item (Generalization)
  For any radius $B_w>0$, we have with high probability (over $(\ab, \Wb_0, \tilde{S}_{n_0})$) that
\begin{align*}
  \E_{(\xb_i, y_i)} \brac{ \sup_{\tfnorm{\Wb\hat\bSigma^{1/2}}\le B_w } \abs{\cR(f^Q_{\Wb}) - \hat{\cR}(f^Q_{\Wb})} } \leq \tilde{O}\left(\frac{B_{g}^2 B_{w}^2 M_{g, \textrm{op}}}{\sqrt{n}} + \frac{1}{\sqrt{n}}\right),
\end{align*}
where $M_{g, \textrm{op}}^2 = B_g^{-2} \EE_{\xb} \left[\opnorm{\frac{1}{n} \sum_{i=1}^n \hb(\xb_i) \hb(\xb_i)^\top}\right]$.
\end{enumerate}
\end{theorem}

\begin{proof}[Proof of Theorem \ref{thm:opt-gen-prime}, Optimization Part]
We recall the second-order directional derivative of $\hat\cR(f^Q_{\Wb})$ satisfies
\begin{align*}
\nabla^2_{\Wb} \hat\cR(f^Q_\Wb)[\Wb_\star, \Wb_\star] \leq \inner{\nabla \hat\cR(f^Q_\Wb)}{\Wb} - 2(\hat\cR(f^Q_\Wb) - \hat\cR(f^Q_{\Wb^*})) + m^{-1} B_g^4\norm{\Wb}_{2, 4}^2 \norm{\Wb_\star}_{2, 4}^2,
\end{align*}
which is established in \eqref{eq:negative-lambda-bound} and $\Wb^*$ is any given matrix. Note we have replaced $B_h$ with $B_g = \norm{\gb(\xb)}_2$, and $B_g$ is upper bounded by $\sqrt{D}$. Similar to the proof \ref{pf:opt}, we specialize $\Wb^*$ to be the optimizer $\Wb^* = \argmin_{\norm{\Wb\hat\bSigma^{1/2}}_{2, 4} \leq B_{w, \star}} \hat{\cR}(f^Q_\Wb)$ and denote its risk $\hat{\cR}(f^Q_{\Wb^*}) = M$. We choose the regularization coefficient as
\begin{equation*}
\lambda = \lambda_0 B_{w, \star}^{-4},
\end{equation*}
where $\lambda_0$ is to be determined. We argue that any second-order stationary point $\hat\Wb$ has to
satisfy $\norm{\hat\Wb\hat\bSigma^{1/2}}_{2, 4} = O(B_{w, \star})$. We already know from proof \ref{pf:opt} that for
any $\Wb$, $\inner{\nabla \hat\cR(f^Q_\Wb)}{\Wb} \geq -2$ holds.

Combining with the fact $$\inner{\nabla_\Wb \left(\norm{\Wb\hat\bSigma^{1/2}}_{2, 4}^4\right)}{\Wb}=4\norm{\Wb\hat\bSigma^{1/2}}_{2, 4}^4,$$ we have simultaneously for all $\Wb$ that
\begin{align*}
\inner{\nabla \hat{\cR}^{\rm data}_{\lambda}(f^Q_\Wb)}{\Wb} & \ge \inner{\nabla_\Wb (\lambda\norm{\Wb \hat\bSigma^{1/2}}_{2, 4}^4)}{\Wb} + \inner{\nabla_\Wb \hat{\cR}(f_\Wb^Q)}{\Wb} \\
& \ge 4\lambda\norm{\Wb\hat\bSigma^{1/2}}_{2, 4}^4 -2.
\end{align*}
Therefore we see that any stationary point $\Wb$ has to satisfy
\begin{align*}
\norm{\Wb\hat\bSigma^{1/2}}_{2, 4} \le (2\lambda)^{-1/4}.
\end{align*}

Choosing
\begin{equation*}
\lambda_0 =\frac{1}{36} (2\tau M + \epsilon),
\end{equation*}
we get
$36 \lambda B_{w, \star}^4=2\tau M + \epsilon$. The second-order directional derivative of $\hat{\cR}^{\rm data}_\lambda(f^Q_{\Wb})$ along direction $\Wb_\star$ is upper bounded by
\begin{align*}
\nabla^2_{\Wb} \hat{\cR}^{\rm data}_\lambda(f^Q_{\Wb})[\Wb_\star, \Wb_\star] & = \nabla^2_{\Wb} \hat{\cR}^{\rm data}(f^Q_{\Wb})[\Wb_\star, \Wb_\star] + \lambda \nabla_{\Wb}^2 \norm{\Wb\hat\bSigma^{1/2}}_{2, 4}^4 [\Wb_\star, \Wb_\star] \\
& = \nabla^2_{\Wb} \hat{\cR}^{\rm data}(f^Q_{\Wb})[\Wb_\star, \Wb_\star] + 4\lambda \sum_{r \leq m}\left(\wb_{\star, r}\hat\bSigma \wb_{\star, r}\right) \left(\wb_{r}\hat\bSigma \wb_{r}\right) \\
& \quad + 8\lambda \sum_{r \leq m} \inner{\wb_r\hat\bSigma^{1/2}}{\wb_{\star, r}\hat\bSigma^{1/2}}^2 \\
& \leq \inner{\nabla \hat\cR(f^Q_\Wb)}{\Wb} - 2(\hat\cR(f^Q_\Wb) - M) + m^{-1} B_g^4\norm{\Wb}_{2, 4}^2 \norm{\Wb_\star}_{2, 4}^2 \\
& \quad + 12 \lambda \norm{\Wb\hat\bSigma^{1/2}}_{2, 4}^2 \norm{\Wb_\star\hat\bSigma^{1/2}}_{2, 4}^2 \\
& \overset{(i)}{\leq} \inner{\nabla \hat\cR(f^Q_\Wb)}{\Wb} - 2(\hat\cR(f^Q_\Wb) - M) + m^{-1} B_g^4\norm{\Wb}_{2, 4}^2 \norm{\Wb_\star}_{2, 4}^2 \\
& \quad + \lambda \norm{\Wb\hat\bSigma^{1/2}}_{2, 4}^4 + 36\lambda\norm{\Wb_\star\hat\bSigma^{1/2}}_{2, 4}^4 \\
& \leq \inner{\nabla \hat\cR_\lambda^{\rm data}(f^Q_\Wb)}{\Wb} - 2(\hat\cR_\lambda^{\rm data}(f^Q_\Wb) - M) \\
& \quad + m^{-1} B_g^4\norm{\Wb}_{2, 4}^2 \norm{\Wb_\star}_{2, 4}^2 - \lambda \norm{\Wb\hat\bSigma^{1/2}}_{2, 4}^4 + 36\lambda \norm{\Wb_\star}_{2, 4}^4. 
\end{align*}
We used the fact $12 ab \leq a^2 + 36b^2$. For a second order-stationary point $\hat\Wb$ of $\hat\cR_\lambda(f^Q_\Wb)$, its gradient vanishes and the Hessian is positive definite. Therefore, we have
\begin{align*}
0 \leq - 2(\hat\cR^{\rm data}_\lambda(f^Q_{\hat\Wb}) - M) + m^{-1} B_g^4\norm{\hat\Wb}_{2, 4}^2 \norm{\Wb_\star}_{2, 4}^2 - \lambda \norm{\hat\Wb\hat\bSigma^{1/2}}_{2, 4}^4 + 36\lambda \norm{\Wb_\star\hat\bSigma^{1/2}}_{2, 4}^4. 
\end{align*}
By Assumption \ref{assumption_covariance}, we have $\lambda_{\min}(\bSigma) \geq \lambda_k$. Moreover, by Lemma \ref{lemma:covariance_relative}, when $n_0 = O\left(\delta^{-2} D \log D\right)$, with probability at least $1-\delta$, we have the following relative concentration of $\hat{\bSigma}$:
\begin{align*}
\frac{1}{2} \bSigma \preceq \hat\bSigma \preceq \frac{3}{2} \bSigma.
\end{align*}
Combining these two ingredients together, we deduce
\begin{align*}
\norm{\Wb}_{2, 4}^4 = \sum_{r=1}^m \norm{\hat\bSigma^{-1/2}\hat\bSigma^{1/2} \wb_r}_2^4 & \leq \sum_{r=1}^m \opnorm{\hat{\bSigma}^{-1/2}}^4 \norm{\hat\bSigma^{1/2}\wb_r}^4 \\
& \leq \sum_{r=1}^m 4\lambda_k^{-2} \norm{\hat\bSigma^{1/2}\wb_r}^4 = 4 \lambda_k^{-2} \tfnorm{\Wb \hat\bSigma}^4.
\end{align*}
Exactly the same argument yields $\tfnorm{\Wb_\star}^4 \leq 4 \lambda_k^{-2} \tfnorm{\Wb_\star \hat\bSigma}^4$. Therefore, we choose $m = 4\epsilon^{-1} \lambda_k^{-2} (2\lambda_0)^{-1/2} C^2 B_g^4 B_{w, \star}^4 \geq \epsilon^{-1}C^2 B_g^4\|\hat\Wb\|_{2, 4}^2 \norm{\Wb_\star}_{2, 4}^2$ and the above inequality implies
\begin{align*}
& 2(\hat\cR_\lambda^{\rm data}(f^Q_{\hat\Wb}) - M) \leq 2\tau M + \epsilon + \epsilon \\
\Longrightarrow ~& \hat\cR_\lambda^{\rm data}(f^Q_{\hat\Wb}) \leq (1+\tau)M + \epsilon.
\end{align*}
Plugging in the naive upper bound $\norm{\gb(\xb)}_2 \leq \sqrt{D}$ in $B_g$, the proof is complete.
\end{proof}

\begin{proof}[Proof of Theorem \ref{thm:opt-gen-prime}, Generalization Part]
Built upon the proof \ref{pf:gen}, we have
\begin{align*}
\EE_{(\xb_i, y_i)} \sbr{\sup_{\norm{\Wb\hat{\bSigma}^{1/2}}_{2, 4} \le B_w} \left|\cR(f^Q_{\Wb}) - \hat{\cR}(f^Q_{\Wb})\right|} \le 2\EE_{(\xb_i, y_i),\boldsymbol{\xi}} \sbr{\sup_{\norm{\Wb\hat{\bSigma}^{1/2}}_{2,4} \le B_w} \left|\frac{1}{n} \sum_{i=1}^n \xi_i \ell(f^Q_\Wb(\xb_i), y_i)\right|},
\end{align*}
where $\xi$ is i.i.d. Rademacher random variables. Recall that the whitened feature is $\hb(\xb) = \hat\bSigma^{-1/2} \gb(\xb)$. We further have
\begin{align*}
& ~\quad \EE_{(\xb_i, y_i), \boldsymbol{\xi}} \sbr{\left|\sup_{\norm{\Wb\hat\bSigma^{1/2}}_{2, 4} \le B_w} \frac{1}{n} \sum_{i=1}^n \xi_i \ell(y_i, f^Q_\Wb(\xb_i))\right|} \\
& \le 4\EE_{\xb_i, \boldsymbol{\xi}}\sbr{\sup_{\norm{\Wb\hat\bSigma^{1/2}}_{2, 4}\le B_w} \frac{1}{\sqrt{m}}\sum_{r\le m} 
    \inner{\frac{1}{n} \sum_{i=1}^n\xi_i a_{r}\phi''(\wb_{0,r}^\top\gb(\xb_i))\hb(\xb_i)\hb(\xb_i)^\top}{\hat\bSigma^{1/2}\wb_r\wb_r^\top\hat\bSigma^{1/2}}} \\
& \quad + \frac{2}{\sqrt{n}} \\
& \le 4\EE_{\xb_i, \boldsymbol{\xi}}\Bigg[\sup_{\norm{\Wb\hat\bSigma^{1/2}}_{2, 4}\le B_w}
    \max_{r\in[m]}\opnorm{\frac{1}{n}\sum_{i=1}^n
    \xi_i\phi''(\wb_{0,r}^\top\hb(\xb_i))\hb(\xb_i)\hb(\xb_i)^\top} \\
& \quad \times
    \frac{1}{\sqrt{m}}\sum_{r\le m}\norm{\hat\bSigma^{1/2}\wb_r\wb_r^\top\hat\bSigma^{1/2}}_{*}\Bigg] +
    \frac{2}{\sqrt{n}} \\
& \le 4\EE_{\xb_i, \boldsymbol{\xi}}\sbr{\max_{r\in[m]}\opnorm{\frac{1}{n}\sum_{i=1}^n
    \xi_i\phi''(\wb_{0,r}^\top \hb(\xb)_i) \hb(\xb_i)\hb(\xb_i^\top)}} \\
& \quad\times \underbrace{\sup_{\norm{\Wb\hat\bSigma^{1/2}}_{2, 4}\le B_w} \frac{1}{\sqrt{m}}\sum_{r\le m}\norm{\hat\bSigma^{1/2}\wb_r}_2^2}_{\le B_w^2} + \frac{2}{\sqrt{n}},
\end{align*}
Consequently, the generalization error is still bounded by
\begin{align*}
\EE_{(\xb_i, y_i)}\sbr{\sup_{\tfnorm{\Wb\hat\bSigma^{1/2}}\le B_w} \left|\cR(f^Q_{\Wb}) - \hat{\cR}(f^Q_{\Wb})\right|} \leq \tilde{O}\left(\frac{B_h^2 B_w^2 M_{h, \textrm{op}}}{\sqrt{n}} \sqrt{\log (Dm)}+ \frac{1}{\sqrt{n}}\right).
\end{align*}
\end{proof}

When using \quadg~to learn low-rank polynomials in the form of $$f_\star(\xb) = \sum_{s=1}^{r_\star} \alpha_s (\bbeta_s^\top \xb)^{p_s} ~~~~\textrm{defined in \eqref{equation:f-star}},$$ we derive the following sample complexity bound.
\begin{theorem}[Sample complexity of \quadg]
Suppose Assumption \ref{assumption_covariance} holds, and there exists some $f_\star$ that achieves low risk: $\cR(f_\star)\le \opt$.
Then for any $\epsilon,\delta \in (0, 1)$ and $\tau = \Theta(1)$, choosing
\begin{align}
  D = \Theta\paren{{\rm poly}(r_\star, p) \sum_s \ltwo{\bbeta_s}^{2\ceil{p_s/2}} \epsilon^{-2}\delta^{-1}},~~~~m \ge \wt{O}\paren{ {\rm poly}(r_\star, D)\epsilon^{-2}\delta^{-1}},
\end{align}
$n_0= \wt{O}(D\delta^{-2})$, and a proper $\lambda > 0$, Algorithm~\ref{alg:unwhiten} achieves the following guarantee:
with probability at least $1-\delta$ over the randomness of data and initialization, any second-order stationary point $\hat{\Wb}$ of $\hat{\cR}^{\rm dreg}_\lambda(f^Q_\Wb)$ satisfies
\begin{align*}
  \cR(f^Q_{\hat{\Wb}}) \le (1+\tau)\opt + \epsilon + \wt{O}\paren{\sqrt{ \frac{{\rm poly}(r_\star, p, \delta^{-1}) \lambda_{\ceil{p/2}}^{-1} \epsilon^{-2}\sum_{s=1}^{r_\star} \ltwo{\bbeta_s}^{2\ceil{p_s/2}}}{n}}}.
\end{align*}
In particular, for any $\epsilon>0$, we can achieve $\cR(f^Q_{\hat{\Wb}}) \le (1+\tau)\opt + 2\epsilon$ with sample complexity
\begin{align}
  \label{equation:sample-complexity}
  n_0 + n \le \wt{O}\paren{ {\rm poly}(r_\star, p, \lambda_{\ceil{p/2}}^{-1}, \epsilon^{-1}, \delta^{-1}) \sum_{s=1}^{r_\star} \ltwo{\bbeta_s}^{2\ceil{p_s/2}} }.
\end{align}
\end{theorem}

\begin{proof}
The proof reproduces that for \quadh~in Sections \ref{bottomexpress}, \ref{topexpress}, and \ref{quadneuralgen}. Specifically, following the same argument in Lemma \ref{lemma:odd}, we can establish the expressivity of \quadh, where for $r = 1, \dots, m_0$, we only need to choose 
\begin{align*}
\wb_r^{s, *} =
\begin{cases}
2\sqrt{\alpha_s} (3r_{\star})^{1/4} m_0^{-1/4} [\boldsymbol{0}^\top, \dots, \ab_s^\top, \dots, \boldsymbol{0}^\top]^\top, & \textrm{for the $s$-th group in~} \cI_1\textrm{~with~}\alpha_s > 0 \\
2\sqrt{|\alpha_s|} (3r_{\star})^{1/4} m_0^{-1/4} [\boldsymbol{0}^\top, \dots, \ab_s^\top, \dots, \boldsymbol{0}^\top]^\top, & \textrm{for the $s$-th group in~} \cI_2\textrm{~with~}\alpha_s < 0 \\
\end{cases}.
\end{align*}
Remember $\cI_1 = \{1, \dots, m/3\}$ where $a_r = 1$ for $r \in \cI_1$ and $\cI_2 = \{m/3+1, 2m/3\}$ with $a_r = -1$. Compared to using whitened representation $\hb(\xb)$, we remove the multiplicative factor $\hat\bSigma^{1/2}$ in $\wb_r^*$ (see Lemma \ref{lemma:topexpress}). The corresponding representation dimension $D = \frac{8\times 50^2 r_{\star}^3 \sum_{s=1}^{r_\star} p_s^5 \norm{\boldsymbol{\beta}_s}_2^{2\lceil p_s/2 \rceil}}{\epsilon^2 \delta}$ and the width $m \geq \frac{54 r_\star D(1+\log \frac{8}{\delta})}{\epsilon^2} \log \frac{1}{\epsilon}$ remain unchanged. Then with probability $1-\delta$, we have
\begin{align*}
\norm{\frac{1}{2\sqrt{m}} \sum_{r=1}^m a_{r} \mathds{1}\{\wb_{0, r}^\top \gb(\xb) \geq 0\} \left((\wb_r^*)^\top \gb(\xb)\right)^2 - f(\xb)}_{L_2} \leq 7 r_\star \epsilon.
\end{align*}
The rest of the proof follows Section \ref{quadneuralgen}, where we need to upper bound $M_{g, \textrm{op}}$, $B_{w, \star}$, and $B_g$, respectively. We use the naive upper bound on $B_g \leq \sqrt{D}$, since each entry of $\gb(\xb)$ is bounded by $1$. By definition, we have
\begin{align*}
B_g^2 M_{g, \textrm{op}}^2 = B_h^2 M_{h, \textrm{op}}^2 \leq 3.
\end{align*}
Lastly, observe $B_{w, \star}^4 = \tfnorm{\Wb^*\hat\bSigma^{1/2}}^4 = \sum_{r=1}^m \norm{\hat\bSigma^{1/2} \wb_r^*}_2^4$. An upper bound has been already derived in Section \ref{quadneuralgen}, which is $108r_\star^2$. As can be seen, quantities $M_{g, \textrm{op}}$, $B_{w, \star}$, and $B_g$ all retain the same order as using the whitened neural representation $\hb(\xb)$ (with possibly different absolute constants). Therefore, in order to achieve
\begin{align*}
\cR(f^Q_{\hat{\Wb}}) & = \cR(f^Q_{\hat{\Wb}}) - \hat\cR(f^Q_{\hat{\Wb}}) + \hat\cR(f^Q_{\hat{\Wb}}) \leq (1+\tau) \mathsf{OPT} + 2\epsilon,
\end{align*}
the sample complexity needs to satisfy
\begin{align*}
n = \tilde{O}\left(\frac{\lambda_{\lceil p/2 \rceil}^{-1} r_{\star}^7}{\epsilon_0^4 \delta^3} \left(\sum_{s=1}^{r_\star} (p_s+1)^5\norm{\boldsymbol{\beta}_s}_2^{2\lceil p_s/2\rceil}\right) \right),
\end{align*}
and $n_0$ stays the same for the covariance estimation. This yields the same sample complexity (again with a potentially different absolute constant) as using the whitened representation $\hb(\xb)$.
\end{proof}

\end{document}